\documentclass{article}

\usepackage{microtype}
\usepackage{graphicx}
\usepackage{booktabs}
\usepackage{multirow}
\usepackage{hyperref}

\usepackage[accepted]{icml2026}

\usepackage{amsmath}
\usepackage{amssymb}
\usepackage{mathtools}
\usepackage{amsthm}

\usepackage[capitalize,noabbrev]{cleveref}

\usepackage{placeins}
\usepackage{float}
\usepackage{stfloats}
\usepackage{longtable}

\icmltitlerunning{Air Quality Arena: A Multi-Region Dataset and Benchmark for Air Quality Forecasting}

\begin{document}

\twocolumn[
 % \icmltitle{AQA-Bench: A Large-Scale Multi-Region Benchmark for Air Quality Forecasting with Time Series Foundation Models}
  \icmltitle{Air Quality Arena: A Large-Scale Multi-Region Ground Monitoring Dataset and Benchmark for Air Quality Forecasting with Time-Series Foundation Models}

  \icmlsetsymbol{equal}{*}

  \begin{icmlauthorlist}
    \icmlauthor{Rishi Bharadwaj}{bits}
    \icmlauthor{Manik Gupta}{bits}
    \icmlauthor{Pandarasamy Arjunan}{iisc}
  \end{icmlauthorlist}

  \icmlaffiliation{bits}{Department of Computer Science and Information Systems, BITS Pilani, Hyderabad Campus, Hyderabad, India}
  \icmlaffiliation{iisc}{Department of Cyber-Physical Systems, Indian Institute of Science, Bengaluru, India}

  \icmlcorrespondingauthor{Pandarasamy Arjunan }{samy@iisc.ac.in}

  \icmlkeywords{air quality forecasting, time series foundation models, benchmark, dataset, climate change}

  \vskip 0.3in
]

\printAffiliationsAndNotice{}

\begin{abstract}
Air pollution causes an estimated 7.9 million premature deaths annually, making accurate forecasting a critical public health priority. Machine learning is increasingly being applied to forecast air pollution levels, yet existing benchmarks remain narrow in both geographic scope and pollutant coverage, and fail to evaluate the latest generation of time series foundation models (TSFMs) on real world, large scale data. We present Air Quality Arena (AQA), a large scale multi-country and multi-pollutant dataset (AQA-Data) and benchmark (AQA-Bench) to address this gap. AQA covers 6 major pollutants over a three year period across 7 diverse countries and 4 continents, with more than 14,000 station-pollutant series, aiming to provide a comprehensive benchmark for air quality tasks. We benchmark this dataset across 11 leading time series foundation models and classical baselines to assess performance on short-term air quality forecasting. Our results demonstrate that TSFMs are effective zero-shot forecasters and consistently outperform classical baselines, with our top-performing model employing a cross-modal architecture that leverages a vision foundation model for time series forecasting. AQA is publicly released at \url{AirQualityArena.github.io}.

\end{abstract}

% ---------------------------------------------------------------
\section{Introduction}
\label{sec:intro}

Air pollution is now the leading environmental cause of premature deaths globally, responsible for an estimated 7.9 million deaths annually~\citep{stateofglobalair2025}, surpassing tobacco as a risk factor.
The \citet{WHO_AirPollution} estimates that 99\% of the global population is exposed to air that exceeds safe pollutant thresholds, underscoring the urgent need for accurate, large scale air quality monitoring and forecasting.
Early forecasting of pollution events can enable timely public health interventions, yet accurate prediction remains difficult due to the complex interplay of meteorological conditions, emission sources, and local geography.

Traditional approaches based on numerical models and statistical methods have seen growing competition from deep learning models. 
More recently, Time Series Foundation Models (TSFMs), large models pretrained on diverse time series corpora that can forecast in a zero-shot setting without task specific training~\citep{ansari2024chronos,das2024timesfm,woo2024moirai}, have come into prominence.
This is particularly appealing for air quality forecasting, where labelled historical data may be sparse, unreliable, or simply unavailable for newly deployed stations, and where retraining a model for each new site or pollutant is impractical at scale. Despite this rapid progress in model development, the benchmarks used to evaluate air quality forecasting have not kept pace.
Most studies evaluate a single country or region \citep{Sharma_2022,silver2025}, focus on one or two pollutants, typically PM$_{2.5}$ \citep{SU2023121074,RAKHOLIA2022101315}, and none systematically compare the new generation of TSFMs against classical baselines at scale across diverse geographies and multiple pollutants.
We introduce AQA to address this gap. 

Rather than relying on aggregated sources such as OpenAQ~\citep{OpenAQ},
which we found failed minimum continuity requirements at almost every
station evaluated, we collected raw measurements directly from six official
national monitoring networks spanning the United States, India, China,
the United Kingdom, Mexico, and France and Germany.
These networks span four continents and 7 countries to capture climatic and emission diversity absent from prior benchmarks, from coal dominated industry in China to agricultural burning in India.

The three year time window (July 2022--June 2025) was chosen to be as recent as possible given the semi-annual update frequency of some sources, and to capture the seasonal variability essential for pollutants such as O$_3$ and PM$_{2.5}$. AQA-Data covers 6 major pollutants (PM$_{2.5}$, PM$_{10}$, NO$_2$, SO$_2$, CO, and O$_3$) comprising more than 14,000 station-pollutant series. We benchmark this dataset across 11 leading TSFMs and 6 classical baselines for short-term air quality forecasting.

\subsection{Contributions}
We present AQA, which, to the best of our knowledge, is the largest and most geographically diverse ML ready air quality dataset and benchmark currently available. It spans regulatory monitoring networks across seven countries, six  pollutants, and supports several tasks including forecasting, classification, and transfer learning. Below are our main contributions:
\begin{itemize}
  \item \textbf{Dataset:} AQA-Data, an air quality dataset spanning
        6 monitoring networks, 7 countries, 4 continents, and 6 pollutants over a
        three-year period (July 2022-June 2025). 
  \item \textbf{Benchmark:} AQA-Bench, a standardised forecasting benchmark evaluated across
        11 time series foundation models and 6 classical baselines for short-term air
        quality prediction.
  \item \textbf{Analysis:} A cross-geography and cross-pollutant evaluation revealing where TSFMs generalise as reliable zero-shot forecasters and where meaningful performance gaps remain, with implications for future model development and benchmark design.
  \item \textbf{Open codebase:} An extensible framework supporting easy integration of
        additional air quality networks, pollutants and models.
\end{itemize}

% ---------------------------------------------------------------
\section{Related Work}
\label{sec:related}

\subsection{Air Quality Datasets}

Several publicly available air quality datasets have been developed to support
environmental monitoring and machine learning research. However, each comes with
notable limitations in terms of temporal coverage, pollutant diversity, geographic
breadth, and suitability for time series forecasting.

\textbf{OpenAQ} aggregates ground-based measurements from government and research networks across more than 100 countries, accessible via a public API~\citep{OpenAQ}.
Its geographic breadth and standardized formatting are appealing, but data quality and continuity vary widely across stations.
It performs no systematic gap filling, leaving preprocessing decisions to
individual researchers. This inconsistency poses a significant obstacle for
reproducible benchmarking.
In our evaluation, OpenAQ data failed to meet minimum continuity requirements at almost every station we assessed, despite drawing from many of the same underlying networks as AQA-Data. For example, based on our analysis, we found data from India with year long gaps. Data is also no longer being collected from China.

\textbf{AQICN} is a similarly broad aggregator, consolidating real time air quality data from thousands of stations worldwide~\citep{AQICN}. However, AQICN currently exposes only calculated AQI indices rather than individual pollutant concentrations, making it unsuitable for multi-pollutant forecasting research.

\textbf{AQ-Bench}~\citep{essd-13-3013-2021} covers over 5,500 stations globally, but is restricted to only ozone and does not include any other pollutants. Its annual aggregated resolution renders it inadequate for time series forecasting tasks.

A consistent pattern emerges from existing datasets. The few that cover multiple pollutants are limited to a single city or country. Those with global reach provide only aggregated statistics or composite indices, with no explicit data preprocessing for building ML pipelines. None address the geographic diversity, pollutant coverage, and pretraining considerations that rigorous TSFM benchmarking demands. AQA is proposed and designed to close all of these gaps simultaneously in a systematic manner.

\subsection{Benchmarking Time Series Models}

Seasonal Naive~\citep{hyndman2018forecasting} and AutoETS~\citep{hyndman2008sm} represent the statistical baselines, simple but competitive on data with strong seasonal structure.
LightGBM~\citep{10.5555/3294996.3295074} is a gradient-boosted tree model trained on lag and calendar features, offering a strong non-neural supervised baseline.
DeepAR~\citep{SALINAS20201181}, DLinear \citep{10.1609/aaai.v37i9.26317} and PatchTST \citep{nie2023a} are supervised deep learning baselines. DeepAR uses an autoregressive recurrent neural network model to produce accurate probabilistic forecasts. DLinear uses simple linear models and has proven to be surprisingly competitive against complex Transformer architectures on standard benchmarks.
PatchTST uses a patch-based Transformer with channel independence, achieving strong results on traffic, weather and electricity forecasting.
Together, these baselines span classical statistics to top performing deep learning methods, providing a
comprehensive reference to evaluate foundation model performance.

TSFMs such as Chronos, Moirai, and TimesFM~\citep{ansari2024chronos,woo2024moirai,das2024timesfm} are pretrained on large, diverse time series corpora and forecast without any task-specific training. 
The closest existing work using TSFMs for air quality forecasting is \citet{saurav2025using}, who evaluate TSFMs on atmospheric CO$_2$ forecasting under zero-shot and fine-tuned settings and assess spatial transfer capabilities across locations.
AQA extends this from a single atmospheric variable to six pollutants across seven countries.
Crucially, our comparison is asymmetric by design, as
TSFMs are evaluated zero-shot while supervised baselines
are fitted per pollutant per network. This directly tests a
practically relevant question, i.e,  can a pretrained foundation
model, without seeing any air quality data, compete with a
model optimised on the target series?

% ---------------------------------------------------------------
\section{AQA-Data}
\label{sec:dataset}

\subsection{Data Sources and Collection}
AQA-Data aggregates ground station measurements from six official air quality monitoring
networks across seven countries. Environmental Protection Agency (EPA) hourly data was
downloaded directly from the Air Quality System (AQS) public data repository~\citep{epa_aqs}.
\citet{cpcb} (CPCB) Observations were collected from the
Indian government's Continuous Ambient Air Quality Monitoring (CAAQM) portal. Automatic Urban and Rural Network
(AURN) data for the United Kingdom \citep{defra_aurn} was retrieved using PyAURN~\citep{pyaurn}, a Python
wrapper around the openair R package~\citep{CARSLAW201252}. Chinese national monitoring data
was collected from \citet{quotsoft}, a third-party aggregator of \citet{cnemc} (CNEMC) station readings. \citet{eea_portal} (EEA) data covering
France and Germany was downloaded directly from the EEA's official data
portal. Mexican data was collected from the Sistema Nacional de
Informaci\'{o}n de la Calidad del Aire (SINAICA) government website~\citep{sinaica}.

To assess potential pretraining contamination, we inspected the publicly available corpora of the evaluated TSFMs. Overlap with AQA-Data was identified only at a small subset of CNEMC stations and AURN stations in London, with the remaining five networks entirely unaffected. For the overlapping stations, we found no temporal intersection between the pretraining data and AQA-Data. Given the limited scale of the overlap and the absence of temporal intersection, we consider contamination effects on the reported results negligible.

\begin{figure}[t]
  \vskip 0.2in
  \centering
  \includegraphics[width=\linewidth]{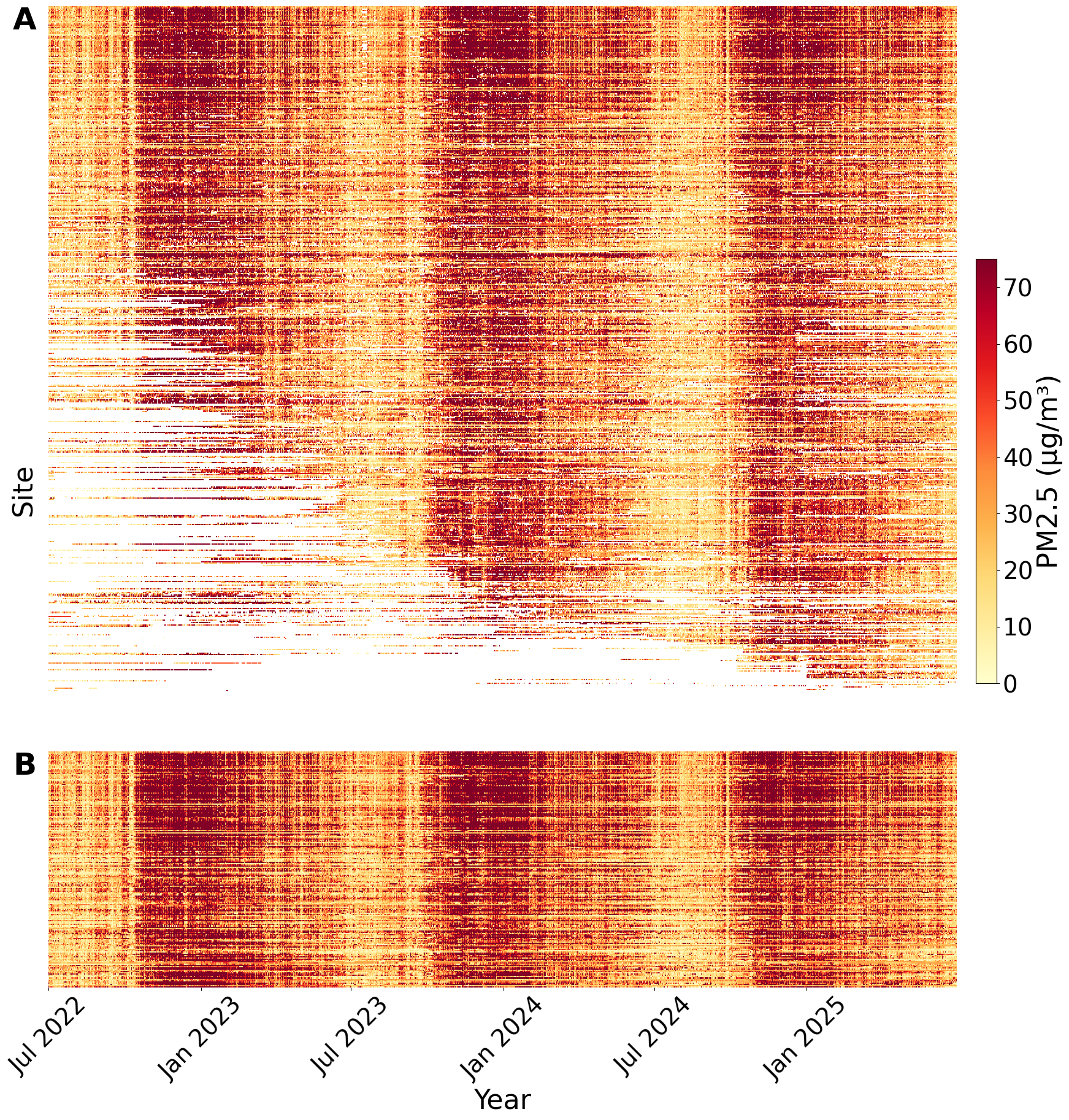}
  \caption{CPCB PM$_{2.5}$ heatmaps across monitoring sites over time, (A) before and (B) after transformation. White regions indicate missing observations. The completeness after transformation reflects removal of sites during filtering and MSTL-based gap imputation.}
  \label{fig:cpcb_heatmaps}
\end{figure}

\subsection{Preprocessing Framework}
All collected data was converted into a uniform format organised by monitoring site and pollutant, covering three years of hourly readings from July 1 2022 to June 30 2025. To balance dataset quantity and quality, we restricted our analysis to monitoring
sites that contained at least 70\% valid observations and had no temporal gaps longer
than two weeks. This filtering was performed separately for each pollutant. Remaining
gaps after site filtering were imputed using Multiple Seasonal-Trend decomposition
using LOESS (MSTL) \citep{cleveland1990}. Interpolating in the deseasonalized space
produces more accurate gap estimates than naive linear interpolation for data with strong seasonality \citep{chhabra2023, wijesekara2023}. Imputed values were clipped
to $[0, x_{\max}]$, where $x_{\max}$ is the maximum observed value for that site and
pollutant, to prevent the introduction of artificial outliers. Figure~\ref{fig:cpcb_heatmaps} shows data from the CPCB network before and after transformation.

% ---------------------------------------------------------------
\section{Benchmark Design}
\label{sec:benchmark}

We adapt the TIME framework~\citep{qiao2026s} as our evaluation
harness, extending it with additional models and training capabilities
for our supervised baselines. All models use a 168-hour context
window and forecast the next 24 hours, evaluated on a rolling window
with a step size of 24. All models operate in a strictly univariate setting
with no covariates. Due
to the large number of stations in CNEMC, we applied spatially
stratified random sampling to select 200 sites per pollutant. Site counts per network and pollutant can be found at Appendix~\ref{app:sites}.
Details regarding model checkpoints, hyperparameters, and training configurations can be found in Appendix~\ref{app:model_information}.
Sites with degenerate Mean Absolute Scaled Error (MASE) or Continuous Ranked Probability Score (CRPS) were excluded from evaluation;
details and excluded site counts are reported in Appendix~\ref{app:excluded_sites}.

\subsection{Results}

\begin{table}[H]
\caption{Pollutant-balanced overall leaderboard}
\label{tab:pollutant_balanced}
\centering
\begin{small}
\begin{tabular}{lrr}
\toprule
model & MASE (norm.) & CRPS (norm.) \\
\midrule
\multicolumn{3}{l}{\textbf{TSFMs}} \\
Chronos-2 & 0.7929 & 0.4654 \\
Chronos-Bolt & 0.7976 & 0.4999 \\
Kairos & 1.0121 & 0.6426 \\
Moirai-1 & 0.8103 & 0.4670 \\
Moirai-2 & 0.7916 & \underline{0.4421} \\
Sundial & 0.7977 & 0.5728 \\
TiRex & \underline{0.7825} & 0.4553 \\
TimesFM-1.0 & 0.8298 & 0.5133 \\
TimesFM-2.0 & 0.8013 & 0.4807 \\
TimesFM-2.5 & \textit{0.7831} & \textbf{0.4385} \\
VisionTS++ & \textbf{0.7785} & \textit{0.4537} \\
\midrule
\multicolumn{3}{l}{\textbf{ML Baselines}} \\
DLinear & 0.8999 & 0.5652 \\
DeepAR & 0.9091 & 0.4911 \\
LightGBM & 0.9268 & 0.5765 \\
PatchTST & 0.8300 & 0.4625 \\
\midrule
\multicolumn{3}{l}{\textbf{Statistical Baselines}} \\
AutoETS & 0.8949 & 0.7878 \\
Seasonal Naive & 1.0000 & 1.0000 \\
\bottomrule
\end{tabular}
\end{small}
\end{table}

Table~\ref{tab:pollutant_balanced} reports the overall leaderboard. Following
\citet{aksu2024gift}, scores are aggregated by
taking the mean over rolling windows, then over series, then over
pollutants, and finally the geometric mean over networks to account for
heterogeneous difficulty levels across datasets. Results are normalized by Seasonal Naive.
Detailed results individually for each network and pollutant before normalization can be found in Appendix~\ref{app:per_pollutant}.

VisionTS++~\citep{shen2025visionts++} achieves the lowest overall MASE, followed by
TiRex~\citep{auer2025tirex} and TimesFM-2.5. VisionTS++ also holds the top spot on each individual country.
Notably, VisionTS++ operates by rendering time series as images and leveraging a vision
foundation model. This cross-modal approach suggests that multimodal pretraining may
confer substantial advantages over purely temporal architectures, and that visual
representations of temporal patterns remain a promising direction for environmental
forecasting.
Among supervised baselines, PatchTST is competitive with lower ranked TSFMs, while DLinear underperforms AutoETS in aggregate, driven by severe degradation on CPCB and SO$_2$ series.
Kairos~\citep{feng2026kairos} is the weakest TSFM overall, the only model to fall below Seasonal Naive. We believe this may be due to its architecture being optimized for much longer context windows than the 168-hour input used here.

Cross-network results (Appendix Table~\ref{tab:per_dataset}) reveal a consistent picture. 
TSFM relative rankings are stable across all seven networks, with VisionTS++, TiRex, and TimesFM-2.5 occupying the top positions on every leaderboard, while absolute forecasting difficulty varies substantially.
AURN and EEA-France are easiest, while CPCB is hardest among well-sampled networks. On CPCB, both DLinear and PatchTST fail to improve on Seasonal Naive despite being trained on that data.
This points to genuine distributional complexity rather than mere domain shift, and confirms the task is hard rather than simply unfamiliar to zero-shot models. Similarly cross-pollutant results follow a consistent hierarchy across networks. Particulate matter is most predictable, followed by NO$_2$ and CO, while SO$_2$ and O$_3$ pose the hardest challenges for all model classes. Notably, DLinear and LightGBM exceed a normalized MASE of 1.0 on SO$_2$.
Within the top tier, VisionTS++ leads on the easier pollutants (CO, NO$_2$,
PM$_{10}$, PM$_{2.5}$) while TiRex leads on O$_3$ and SO$_2$.

The networks where zero-shot performance is weakest,
CPCB and SINAICA, are also the ones where forecasting failures carry the greatest consequence. Indian stations
in AQA-Bench record mean PM$_{2.5}$ concentrations of 57
$\mu$g/m$^3$ and PM$_{10}$ of 122 $\mu$g/m$^3$ (Table~\ref{tab:data_statistics}), roughly twelve and
eight times the WHO annual guidelines respectively~\citep{WHO2021AQG}. A model benchmarked only
on European or US data would appear to work well while
remaining unvalidated precisely where pollution is worst.

\section{Conclusion and Future Work}

AQA-Bench establishes TSFMs as a viable and broadly applicable approach to air quality forecasting, outperforming statistical baselines across geographies and pollutants in a zero-shot setting.
Our work answers a practically important question, showing that a pretrained foundation model can outperform models optimised directly on the target series.
The performance differences observed across urban profiles highlight why benchmark diversity matters.
Evaluating only on cleaner networks obscures how models perform precisely where pollution is highest and the consequences of forecasting failures are greatest.
AQA-Bench addresses these gaps and provides a more representative evaluation framework for diverse air quality conditions.

The gaps AQA-Bench surfaces motivate important directions for future work. 
Expanding geographic coverage, few-shot adaptation to high-pollution networks, evaluating multiple horizons, and modelling additional covariates for reactive pollutants are all directions we hope to pursue. In particular, exploration of cross-modal architectures following the strong performance of VisionTS++ seems to be a very promising direction to pursue. We encourage the community to do the same 
using our open-sourced framework.

% ---------------------------------------------------------------

\bibliographystyle{icml2026}
\bibliography{references}

@article{Sharma_2022, title={Analysis of Air Pollution Data in India between 2015 and 2019}, volume={22}, ISSN={2071-1409}, url={http://dx.doi.org/10.4209/aaqr.210204}, DOI={10.4209/aaqr.210204}, number={2}, journal={Aerosol and Air Quality Research}, publisher={Springer Science and Business Media LLC}, author={Sharma, Disha and Mauzerall, Denise}, year={2022}, pages={210204} }

@article{silver2025,
  title   = {A decade of {China}'s air quality monitoring data suggests
             health impacts are no longer declining},
  author  = {Silver, Ben and Reddington, Carly L. and Chen, Yue and Arnold, Steve R.},
  journal = {Environment International},
  volume  = {197},
  pages   = {109318},
  year    = {2025},
  doi     = {10.1016/j.envint.2025.109318}
}

@inproceedings{saurav2025using,
  title={Using Time Series Foundation Models for Atmospheric CO2 Concentration Forecasting},
  author={Saurav, Kumar and Baghel, Vinamra and Jain, Ayush and Guruprasad, Ranjini},
  booktitle={NeurIPS 2025 Workshop on Tackling Climate Change with Machine Learning},
  url={https://www.climatechange.ai/papers/neurips2025/76},
  year={2025}
}

@misc{epa_aqs,
  title        = {{EPA} Air Quality System ({AQS}) Data},
  author       = {{U.S. Environmental Protection Agency}},
  url          = {https://aqs.epa.gov/aqsweb/airdata/download_files.html#Raw}
}

@misc{cpcb,
  title        = {Continuous Ambient Air Quality Monitoring ({CAAQM}) Data Repository},
  author       = {{Central Pollution Control Board}},
  url          = {https://airquality.cpcb.gov.in/ccr/#/caaqm-dashboard-all/caaqm-landing/caaqm-data-repository}
}

@article{CARSLAW201252,
title = {openair — An R package for air quality data analysis},
journal = {Environmental Modelling \& Software},
volume = {27-28},
pages = {52-61},
year = {2012},
issn = {1364-8152},
doi = {https://doi.org/10.1016/j.envsoft.2011.09.008},
url = {https://www.sciencedirect.com/science/article/pii/S1364815211002064},
author = {David C. Carslaw and Karl Ropkins}
}

@misc{quotsoft,
  title        = {China Air Quality Historical Data},
  author       = {{quotsoft.net}},
  url          = {https://quotsoft.net/air/}
}

@misc{eea_portal,
  title        = {Air Quality Download Service},
  author       = {{European Environment Agency}},
  url          = {https://www.eea.europa.eu/en/datahub/datahubitem-view/778ef9f5-6293-4846-badd-56a29c70880d}
}

@misc{sinaica,
  title        = {{SINAICA} --- Sistema Nacional de Información de la Calidad del Aire},
  author       = {{Gobierno de México}},
  url          = {https://sinaica.inecc.gob.mx/}
}

@misc{pyaurn,
  title        = {{PyAURN}: A Python wrapper for accessing {AURN} air quality data},
  author       = {Wilson, Robin},
  url          = {https://github.com/robintw/PyAURN}
}

@misc{defra_aurn,
  author       = {{Department for Environment, Food \& Rural Affairs}},
  title        = {Automatic Urban and Rural Network ({AURN})},
  howpublished = {\url{https://uk-air.defra.gov.uk/networks/network-info?view=aurn}},
  note         = {Accessed: 2026-04-29}
}

@misc{cnemc,
  author       = {{China National Environmental Monitoring Centre}},
  title        = {{China National Environmental Monitoring Centre (CNEMC)}},
  howpublished = {\url{https://www.cnemc.cn/en/}},
  note         = {Established 1980; Accessed: 2026-04-29}
}

@article{chhabra2023,
  title   = {Comparison of Imputation Methods for Univariate Time Series},
  author  = {Chhabra, G.},
  journal = {International Journal on Recent and Innovation Trends in Computing and Communication},
  volume  = {11},
  number  = {2s},
  pages   = {286--292},
  year    = {2023},
  doi     = {10.17762/ijritcc.v11i2s.6148}
}

@article{wijesekara2023,
  title   = {Mind the Large Gap: Novel Algorithm Using Seasonal Decomposition and 
             Elastic Net Regression to Impute Large Intervals of Missing Data in 
             Air Quality Data},
  author  = {Wijesekara, L. and Liyanage, L.},
  journal = {Atmosphere},
  volume  = {14},
  number  = {2},
  pages   = {355},
  year    = {2023},
  doi     = {10.3390/atmos14020355}
}

@article{cleveland1990,
  author    = {Cleveland, Robert B. and Cleveland, William S. and McRae, Jean E. and Terpenning, Irma},
  title     = {{STL}: A Seasonal-Trend Decomposition Procedure Based on {Loess}},
  journal   = {Journal of Official Statistics},
  volume    = {6},
  number    = {1},
  pages     = {3--73},
  year      = {1990},
  publisher = {Statistics Sweden}
}

@misc{garza2022statsforecast,
    author={Azul Garza and Max Mergenthaler Canseco and Cristian Challú and Kin G. Olivares},
    title = {{StatsForecast}: Lightning fast forecasting with statistical and econometric models},
    year={2022},
    howpublished={{PyCon} Salt Lake City, Utah, US 2022},
    url={https://github.com/Nixtla/statsforecast}
}

@misc{aksu2024gift,
      title={GIFT-Eval: A Benchmark For General Time Series Forecasting Model Evaluation}, 
      author={Taha Aksu and Gerald Woo and Juncheng Liu and Xu Liu and Chenghao Liu and Silvio Savarese and Caiming Xiong and Doyen Sahoo},
      year={2024},
      eprint={2410.10393},
      archivePrefix={arXiv},
      primaryClass={cs.LG},
      url={https://arxiv.org/abs/2410.10393}, 
}

@article{qiao2026s,
  title={It's TIME: Towards the Next Generation of Time Series Forecasting Benchmarks},
  author={Qiao, Zhongzheng and Pan, Sheng and Wang, Anni and Zhukova, Viktoriya and Liu, Yong and Jiang, Xudong and Wen, Qingsong and Long, Mingsheng and Jin, Ming and Liu, Chenghao},
  journal={arXiv preprint arXiv:2602.12147},
  year={2026}
}

@misc{OpenAQ,
  author       = {{OpenAQ}},
  title        = {OpenAQ},
  howpublished = {\url{https://openaq.org/} (accessed March 17, 2026)},
}

@misc{AQICN,
  author       = {{World Air Quality Index Project}},
  title        = {World Air Quality Index},
  howpublished = {\url{https://aqicn.org/} (accessed April 17, 2026)},
}

@Article{essd-13-3013-2021,
AUTHOR = {Betancourt, C. and Stomberg, T. and Roscher, R. and Schultz, M. G. and Stadtler, S.},
TITLE = {AQ-Bench: a benchmark dataset for machine learning on global air quality metrics},
JOURNAL = {Earth System Science Data},
VOLUME = {13},
YEAR = {2021},
NUMBER = {6},
PAGES = {3013--3033},
URL = {https://essd.copernicus.org/articles/13/3013/2021/},
DOI = {10.5194/essd-13-3013-2021}
}

@article{liu2025moirai2,
  title={Moirai 2.0: When less is more for time series forecasting},
  author={Liu, Chenghao and Aksu, Taha and Liu, Juncheng and Liu, Xu and Yan, Hanshu and Pham, Quang and Sahoo, Doyen and Xiong, Caiming and Savarese, Silvio and Li, Junnan},
  journal={arXiv preprint arXiv:2511.11698},
  year={2025}
}

@inproceedings{woo2024moirai,
  title={Unified Training of Universal Time Series Forecasting Transformers},
  author={Woo, Gerald and Liu, Chenghao and Kumar, Akshat and Xiong, Caiming and Savarese, Silvio and Sahoo, Doyen},
  booktitle={Forty-first International Conference on Machine Learning},
  year={2024}
}

@article{ansari2024chronos,
  title={Chronos: Learning the Language of Time Series},
  author={Ansari, Abdul Fatir and Stella, Lorenzo and Turkmen, Caner and Zhang, Xiyuan and Mercado, Pedro and Shen, Huibin and Shchur, Oleksandr and Rangapuram, Syama Syndar and Pineda Arango, Sebastian and Kapoor, Shubham and Zschiegner, Jasper and Maddix, Danielle C. and Mahoney, Michael W. and Torkkola, Kari and Gordon Wilson, Andrew and Bohlke-Schneider, Michael and Wang, Yuyang},
  journal={Transactions on Machine Learning Research},
  issn={2835-8856},
  year={2024},
  url={https://openreview.net/forum?id=gerNCVqqtR}
}

@article{ansari2025chronos2,
  title={Chronos-2: From univariate to universal forecasting},
  author={Ansari, Abdul Fatir and Shchur, Oleksandr and K{\"u}ken, Jaris and Auer, Andreas and Han, Boran and Mercado, Pedro and Rangapuram, Syama Sundar and Shen, Huibin and Stella, Lorenzo and Zhang, Xiyuan and others},
  journal={arXiv preprint arXiv:2510.15821},
  year={2025}
}

@article{shen2025visionts++,
  title={VisionTS++: Cross-Modal Time Series Foundation Model with Continual Pre-trained Vision Backbones},
  author={Shen, Lefei and Chen, Mouxiang and Liu, Xu and Fu, Han and Ren, Xiaoxue and Sun, Jianling and Li, Zhuo and Liu, Chenghao},
  journal={arXiv preprint arXiv:2508.04379},
  year={2025}
}

@inproceedings{auer2025tirex,
title={TiRex: Zero-Shot Forecasting Across Long and Short Horizons with Enhanced In-Context Learning},
author={Andreas Auer and Patrick Podest and Daniel Klotz and Sebastian B{\"o}ck and G{\"u}nter Klambauer and Sepp Hochreiter},
booktitle={The Thirty-ninth Annual Conference on Neural Information Processing Systems},
year={2025},
url={https://openreview.net/forum?id=v7UqniC9pF}
}

@inproceedings{liu2025sundial,
title={Sundial: A Family of Highly Capable Time Series Foundation Models},
author={Yong Liu and Guo Qin and Zhiyuan Shi and Zhi Chen and Caiyin Yang and Xiangdong Huang and Jianmin Wang and Mingsheng Long},
booktitle={Forty-second International Conference on Machine Learning},
year={2025},
url={https://openreview.net/forum?id=LO7ciRpjI5}
}

@misc{feng2026kairos,
      title={Kairos: Toward Adaptive and Parameter-Efficient Time Series Foundation Models}, 
      author={Kun Feng and Shaocheng Lan and Yuchen Fang and Wenchao He and Sihan Lu and Shuqi Gu and Lintao Ma and Xingyu Lu and Kan Ren},
      year={2026},
      eprint={2509.25826},
      archivePrefix={arXiv},
      primaryClass={cs.LG},
      url={https://arxiv.org/abs/2509.25826}, 
}

@misc{das2024timesfm,
      title={A decoder-only foundation model for time-series forecasting}, 
      author={Abhimanyu Das and Weihao Kong and Rajat Sen and Yichen Zhou},
      year={2024},
      eprint={2310.10688},
      archivePrefix={arXiv},
      primaryClass={cs.CL},
      url={https://arxiv.org/abs/2310.10688}, 
}

@techreport{stateofglobalair2025,
  author      = {{Health Effects Institute}},
  title       = {State of Global Air 2025: A Report on Air Pollution and Its Role in the World's Leading Causes of Death},
  year        = {2025},
  institution = {Health Effects Institute},
  address     = {Boston, MA},
  url         = {https://www.stateofglobalair.org},
  note        = {Accessed: April 20, 2026}
}

@book{WHO2021AQG,
  author    = {{World Health Organization}},
  title     = {{WHO} Global Air Quality Guidelines: Particulate Matter ({PM2.5} and {PM10}), Ozone, Nitrogen Dioxide, Sulfur Dioxide and Carbon Monoxide},
  year      = {2021},
  publisher = {World Health Organization},
  address   = {Geneva},
  note      = {Licence: CC BY-NC-SA 3.0 IGO}
}

@misc{WHO_AirPollution,
  author       = {{World Health Organization}},
  title        = {Air Pollution},
  howpublished = {\url{https://www.who.int/health-topics/air-pollution}},
  note         = {Accessed: 2026-04-21}
}

@article{SU2023121074,
title = {Effective PM2.5 concentration forecasting based on multiple spatial-temporal GNN for areas without monitoring stations},
journal = {Expert Systems with Applications},
volume = {234},
pages = {121074},
year = {2023},
issn = {0957-4174},
doi = {https://doi.org/10.1016/j.eswa.2023.121074},
url = {https://www.sciencedirect.com/science/article/pii/S0957417423015762},
author = {I-Fang Su and Yu-Chi Chung and Chiang Lee and Pin-Man Huang}
}

@article{RAKHOLIA2022101315,
title = {AI-based air quality PM2.5 forecasting models for developing countries: A case study of Ho Chi Minh City, Vietnam},
journal = {Urban Climate},
volume = {46},
pages = {101315},
year = {2022},
issn = {2212-0955},
doi = {https://doi.org/10.1016/j.uclim.2022.101315},
url = {https://www.sciencedirect.com/science/article/pii/S2212095522002334},
author = {Rajnish Rakholia and Quan Le and Khue Vu and Bang Quoc Ho and Ricardo Simon Carbajo}
}

@inproceedings{10.1609/aaai.v37i9.26317,
author = {Zeng, Ailing and Chen, Muxi and Zhang, Lei and Xu, Qiang},
title = {Are transformers effective for time series forecasting?},
year = {2023},
isbn = {978-1-57735-880-0},
publisher = {AAAI Press},
url = {https://doi.org/10.1609/aaai.v37i9.26317},
doi = {10.1609/aaai.v37i9.26317},
booktitle = {Proceedings of the Thirty-Seventh AAAI Conference on Artificial Intelligence and Thirty-Fifth Conference on Innovative Applications of Artificial Intelligence and Thirteenth Symposium on Educational Advances in Artificial Intelligence},
articleno = {1248},
numpages = {8},
series = {AAAI'23/IAAI'23/EAAI'23}
}

@inproceedings{nie2023a,
title={A Time Series is Worth 64 Words:  Long-term Forecasting with Transformers},
author={Yuqi Nie and Nam H Nguyen and Phanwadee Sinthong and Jayant Kalagnanam},
booktitle={The Eleventh International Conference on Learning Representations },
year={2023},
url={https://openreview.net/forum?id=Jbdc0vTOcol}
}

@book{hyndman2018forecasting,
title = "Forecasting: Principles and Practice",
author = "Hyndman, \{Robin John\} and George Athanasopoulos",
year = "2018",
language = "English",
publisher = "OTexts",
address = "Australia"
}

@book{hyndman2008sm,
  title={Forecasting with Exponential Smoothing: The State Space Approach},
  author={Hyndman, R. and Koehler, A.B. and Ord, J.K. and Snyder, R.D.},
  isbn={9783540719182},
  series={Springer Series in Statistics},
  url={https://books.google.co.in/books?id=GSyzox8Lu9YC},
  year={2008},
  publisher={Springer Berlin Heidelberg}
}

@article{agtabular,
  title={AutoGluon-Tabular: Robust and Accurate AutoML for Structured Data},
  author={Erickson, Nick and Mueller, Jonas and Shirkov, Alexander and Zhang, Hang and Larroy, Pedro and Li, Mu and Smola, Alexander},
  journal={arXiv preprint arXiv:2003.06505},
  year={2020}
}

@inproceedings{agtimeseries,
  title={{AutoGluon-TimeSeries}: {AutoML} for Probabilistic Time Series Forecasting},
  author={Shchur, Oleksandr and Turkmen, Caner and Erickson, Nick and Shen, Huibin and Shirkov, Alexander and Hu, Tony and Wang, Yuyang},
  booktitle={International Conference on Automated Machine Learning},
  year={2023}
}

@inproceedings{10.5555/3294996.3295074,
author = {Ke, Guolin and Meng, Qi and Finley, Thomas and Wang, Taifeng and Chen, Wei and Ma, Weidong and Ye, Qiwei and Liu, Tie-Yan},
title = {LightGBM: a highly efficient gradient boosting decision tree},
year = {2017},
isbn = {9781510860964},
publisher = {Curran Associates Inc.},
address = {Red Hook, NY, USA},
booktitle = {Proceedings of the 31st International Conference on Neural Information Processing Systems},
pages = {3149-3157},
numpages = {9},
location = {Long Beach, California, USA},
series = {NIPS'17}
}

@article{SALINAS20201181,
title = {DeepAR: Probabilistic forecasting with autoregressive recurrent networks},
journal = {International Journal of Forecasting},
volume = {36},
number = {3},
pages = {1181-1191},
year = {2020},
issn = {0169-2070},
doi = {https://doi.org/10.1016/j.ijforecast.2019.07.001},
url = {https://www.sciencedirect.com/science/article/pii/S0169207019301888},
author = {David Salinas and Valentin Flunkert and Jan Gasthaus and Tim Januschowski}
}
% ---------------------------------------------------------------
\section*{Acknowledgements}
We would like to acknowledge all government entities who made 
this data public, as well as quotsoft.net for collecting the 
historical CNEMC data.  We would also like to acknowledge the 
Anuradha and Prashanth Palakurthi Centre for Artificial 
Intelligence Research (APPCAIR), BITS Pilani for funding support.

\newpage
\appendix

\section {Monitoring Sites per Pollutant and Network}
\label{app:sites}
\begin{table}[h]
\centering
\caption{Constituent datasets in AQA-Data.}
\label{tab:datasets}
\begin{small}
\begin{tabular}{p{1.25cm}p{2.2cm}p{3.2cm}}
\toprule
\textbf{Network} & \textbf{Country} & \textbf{Source} \\
\midrule
EPA AQS   & United States & \citet{epa_aqs} \\
CPCB      & India         & \citet{cpcb} \\
AURN      & United Kingdom & \citet{defra_aurn} \\
CNEMC     & China         & \citet{quotsoft} \\
EEA       & France, Germany & \citet{eea_portal} \\
SINAICA   & Mexico        & \citet{sinaica} \\
\bottomrule
\end{tabular}
\end{small}
\end{table}

All data collected was present at an hourly temporal resolution, except CPCB data. CPCB data was collected at a 15 minute resolution and resampled to hourly via median aggregation. After quality filtering (requiring at least 70\% valid observations and no gaps exceeding two weeks), the number of retained stations varies substantially across networks and pollutants. CNEMC has by far the largest coverage; for evaluation it was subsampled to 200 sites per pollutant via spatially stratified random sampling. CO coverage is sparse for AURN and EEA-FR, therefore results should be interpreted with caution.

\begin{table}[H]
  \caption{Number of monitoring sites per pollutant and network in AQA-Data.}
  \label{tab:sites}
  \begin{center}
    \begin{small}
      \begin{tabular}{lrrrrrr}
        \toprule
        Network   & CO & NO$_2$ & O$_3$ & PM$_{10}$ & PM$_{2.5}$ & SO$_2$ \\
        \midrule
        EPA       & 90   & 212 & 488  & 220  & 445  & 212 \\
        AURN
                  & 1    & 70  & 39   & 80   & 61   & 8   \\
        CPCB      & 179  & 179 & 182  & 191  & 190  & 183 \\
        CNEMC     & 1487 & 1481& 1488 & 1481 & 1471 & 1485\\
        EEA-FR    & 5    & 219 & 211  & 204  & 131  & 37  \\
        EEA-DE    & 65   & 370 & 262  & 316  & 246  & 75  \\
        SINAICA   & 12   & 11  & 15   & 16   & 6    & 15  \\
        \midrule
        \multicolumn{7}{c}{\textbf{Total: 14,139}} \\
        \bottomrule
      \end{tabular}
    \end{small}
  \end{center}
\end{table}

\section{Model Information}
\label{app:model_information}

All experiments were conducted on a single NVIDIA RTX A5000 GPU. We used the following checkpoint variants:
Moirai (Base)~\citep{woo2024moirai}, Moirai-2 (Small)~\citep{liu2025moirai2},
Chronos-Bolt (Base) and Chronos-2 (Base)~\citep{ansari2024chronos,ansari2025chronos2},
Kairos (50M)~\citep{feng2026kairos}, Sundial (Base, 128M)~\citep{liu2025sundial},
VisionTS++ (Base)~\citep{shen2025visionts++}, TimesFM-1.0 (Base, 200M),
2.0 (Base, 500M), and 2.5 (Base, 200M)~\citep{das2024timesfm}, and TiRex
(Base)~\citep{auer2025tirex}.
Seasonal Naive and AutoETS were used from the statsforecast library~\citep{garza2022statsforecast}. DeepAR, LightGBM, DLinear, and PatchTST were implemented from the AutoGluon library~\citep{agtabular,agtimeseries}.

Data is split chronologically, the first year of
data serves as the training split and the remaining two years as the
evaluation split. As the majority of models require no training, a larger test split better captures zero-shot performance across seasonal variation.
All evaluated TSFMs support native probabilistic
forecasting and were used without any weight updates or adaptation to
AQA-Data. ML baselines were trained per dataset per pollutant on the training split and evaluated on the test split. Statistical baselines are fitted directly on each test window, as is standard for this model class.

AutoETS was set to AZA (additive errors, automatic trend, additive seasonality) to prevent multiplicative zero errors arising from near-zero pollutant values.
DeepAR, PatchTST and DLinear were trained for 100 epochs, with an early stopping patience of 10. LightGBM was trained for 1000 boost rounds, with the same early stopping patience.

\section{Excluded Sites}
\label{app:excluded_sites}
MASE is computed using a daily seasonal naive baseline (lag 24) over each 168-hour context window.
Sites with a mean MASE or CRPS over all models exceeding 50 were
excluded as degenerate, typically arising from near-constant or
near-zero series where the Seasonal Naive denominator collapses.
Exclusion was applied independently per pollutant and primarily
affected SO$_2$ in the EPA and EEA-DE datasets. Exact excluded site
counts are reported below. No exclusion was
applied for MAE or RMSE.

\begin{table}[H]
  \caption{Sites excluded per network and pollutant due to degenerate
    MASE or CRPS scores (mean $>50$ across all models).}
  \label{tab:excluded_sites}
  \begin{center}
    \begin{small}
\begin{tabular}{lrrr r}
      \toprule
      Network & CO & NO$_2$ & SO$_2$ & Total \\
      \midrule
      CPCB    & 1  & --     & --     & 1  \\
      EEA-DE  & -- & 2      & 20     & 22 \\
      EPA     & 2  & --     & 53     & 55 \\
      \midrule
      \textbf{All others} & -- & -- & -- & 0 \\
      \bottomrule
    \end{tabular}
\end{small}
  \end{center}
\end{table}

\section{Per-Pollutant Results}
\label{app:per_pollutant}

Results are organised by monitoring network (alphabetical), with all six pollutants
shown per network.
\FloatBarrier
\subsection{AURN}
\begin{table}[H]
\caption{CO leaderboard --- AURN}
\centering
\begin{small}
\begin{tabular}{lrrrr}
\toprule
model & MASE & CRPS & MAE & RMSE \\
\midrule
\multicolumn{5}{l}{\textbf{TSFMs}} \\
Chronos-2 & 0.8234 & 0.2082 & 0.0491 & 0.0886 \\
Chronos-Bolt & 0.8380 & 0.2113 & 0.0500 & 0.0914 \\
Kairos & 0.9203 & 0.2376 & 0.0558 & 0.0982 \\
Moirai-1 & 0.8451 & 0.2155 & 0.0502 & 0.0910 \\
Moirai-2 & 0.7988 & 0.2011 & 0.0473 & \textit{0.0854} \\
Sundial & 0.8210 & 0.2241 & 0.0492 & 0.0881 \\
TiRex & \underline{0.7932} & 0.2003 & \underline{0.0467} & \underline{0.0850} \\
TimesFM-1.0 & 0.8306 & 0.2090 & 0.0488 & 0.0858 \\
TimesFM-2.0 & 0.8168 & 0.2068 & 0.0482 & 0.0870 \\
TimesFM-2.5 & 0.7980 & \textit{0.2003} & 0.0473 & 0.0857 \\
VisionTS++ & \textbf{0.7881} & \underline{0.1983} & \textbf{0.0465} & \textbf{0.0849} \\
\midrule
\multicolumn{5}{l}{\textbf{ML Baselines}} \\
DLinear & 0.8928 & 0.2312 & 0.0545 & 0.0953 \\
DeepAR & 0.8960 & 0.2297 & 0.0546 & 0.1005 \\
LightGBM & 1.0031 & 0.2411 & 0.0588 & 0.1048 \\
PatchTST & \textit{0.7951} & \textbf{0.1939} & \textit{0.0472} & 0.0913 \\
\midrule
\multicolumn{5}{l}{\textbf{Statistical Baselines}} \\
AutoETS & 0.9436 & 0.2508 & 0.0571 & 0.0967 \\
Seasonal Naive & 1.0884 & 0.3096 & 0.0667 & 0.1222 \\
\bottomrule
\end{tabular}
\end{small}
\end{table}
\begin{table}[H]
\caption{NO2 leaderboard --- AURN}
\centering
\begin{small}
\begin{tabular}{lrrrr}
\toprule
model & MASE & CRPS & MAE & RMSE \\
\midrule
\multicolumn{5}{l}{\textbf{TSFMs}} \\
Chronos-2 & 0.8363 & 0.3816 & 7.1325 & 10.5645 \\
Chronos-Bolt & 0.8256 & 0.3810 & \textit{7.0254} & 10.3525 \\
Kairos & 1.0233 & 0.4626 & 8.9212 & 12.4706 \\
Moirai-1 & 0.8514 & 0.3881 & 7.2730 & 10.6402 \\
Moirai-2 & 0.8323 & 0.3741 & 7.0968 & 10.4744 \\
Sundial & 0.8341 & 0.4260 & 7.1247 & 10.3836 \\
TiRex & 0.8319 & 0.3798 & 7.1207 & 10.5456 \\
TimesFM-1.0 & 0.8432 & 0.3850 & 7.1944 & 10.6455 \\
TimesFM-2.0 & 0.8297 & 0.3724 & 7.0633 & 10.5288 \\
TimesFM-2.5 & \textit{0.8242} & \textit{0.3688} & 7.0258 & 10.4115 \\
VisionTS++ & \textbf{0.8143} & \underline{0.3684} & \underline{6.9311} & \textit{10.2834} \\
\midrule
\multicolumn{5}{l}{\textbf{ML Baselines}} \\
DLinear & 0.8282 & 0.3814 & 7.0337 & \underline{10.2002} \\
DeepAR & 0.8524 & \textbf{0.3643} & 7.1747 & 10.8434 \\
LightGBM & 0.8507 & 0.3896 & 7.2321 & 10.6544 \\
PatchTST & \underline{0.8156} & 0.3697 & \textbf{6.9258} & \textbf{10.1628} \\
\midrule
\multicolumn{5}{l}{\textbf{Statistical Baselines}} \\
AutoETS & 0.9368 & 0.4591 & 7.9941 & 11.1956 \\
Seasonal Naive & 1.0634 & 0.5459 & 9.1065 & 13.1011 \\
\bottomrule
\end{tabular}
\end{small}
\end{table}
\begin{table}[H]
\caption{Ozone leaderboard --- AURN}
\centering
\begin{small}
\begin{tabular}{lrrrr}
\toprule
model & MASE & CRPS & MAE & RMSE \\
\midrule
\multicolumn{5}{l}{\textbf{TSFMs}} \\
Chronos-2 & 0.8091 & 0.2173 & 11.8232 & 15.9042 \\
Chronos-Bolt & 0.8135 & 0.2182 & 11.8944 & 15.9953 \\
Kairos & 1.0052 & 0.2715 & 14.6526 & 19.3464 \\
Moirai-1 & 0.8310 & 0.2235 & 12.1426 & 16.1774 \\
Moirai-2 & 0.8133 & 0.2198 & 11.9039 & 15.9355 \\
Sundial & 0.7964 & 0.2378 & 11.6216 & \textit{15.4175} \\
TiRex & \textbf{0.7799} & \textbf{0.2116} & \textbf{11.3724} & \textbf{15.3199} \\
TimesFM-1.0 & 0.8578 & 0.2421 & 12.5846 & 16.7826 \\
TimesFM-2.0 & 0.7953 & 0.2201 & 11.6071 & 15.5203 \\
TimesFM-2.5 & \underline{0.7866} & \textit{0.2152} & \underline{11.4601} & \underline{15.3572} \\
VisionTS++ & \textit{0.7930} & \underline{0.2132} & \textit{11.5670} & 15.4976 \\
\midrule
\multicolumn{5}{l}{\textbf{ML Baselines}} \\
DLinear & 0.8048 & 0.2328 & 11.7243 & 15.6214 \\
DeepAR & 0.8175 & 0.2633 & 11.9502 & 15.8906 \\
LightGBM & 0.8591 & 0.2600 & 12.5791 & 16.3401 \\
PatchTST & 0.8153 & 0.2414 & 11.8864 & 15.8026 \\
\midrule
\multicolumn{5}{l}{\textbf{Statistical Baselines}} \\
AutoETS & 0.8976 & 0.2511 & 13.1541 & 17.1105 \\
Seasonal Naive & 1.0503 & 0.2999 & 15.4741 & 20.3091 \\
\bottomrule
\end{tabular}
\end{small}
\end{table}
\begin{table}[H]
\caption{PM10 leaderboard --- AURN}
\centering
\begin{small}
\begin{tabular}{lrrrr}
\toprule
model & MASE & CRPS & MAE & RMSE \\
\midrule
\multicolumn{5}{l}{\textbf{TSFMs}} \\
Chronos-2 & 0.8297 & 0.3434 & 5.2985 & 8.3368 \\
Chronos-Bolt & 0.8361 & 0.3483 & 5.3399 & 8.4259 \\
Kairos & 0.8857 & 0.3737 & 5.7243 & 8.8172 \\
Moirai-1 & 0.8337 & 0.3488 & 5.3336 & 8.4093 \\
Moirai-2 & 0.8269 & \underline{0.3346} & 5.2644 & 8.3249 \\
Sundial & 0.8229 & 0.3732 & 5.2762 & 8.2389 \\
TiRex & 0.8182 & 0.3396 & 5.2249 & 8.3025 \\
TimesFM-1.0 & 0.8462 & 0.3551 & 5.4182 & 8.4812 \\
TimesFM-2.0 & 0.8315 & 0.3464 & 5.3194 & 8.3669 \\
TimesFM-2.5 & \textit{0.8149} & \textbf{0.3326} & \textit{5.1859} & 8.2257 \\
VisionTS++ & \underline{0.8107} & \textit{0.3357} & \underline{5.1600} & \textit{8.1505} \\
\midrule
\multicolumn{5}{l}{\textbf{ML Baselines}} \\
DLinear & 0.8176 & 0.3497 & 5.2826 & \underline{8.1483} \\
DeepAR & 0.8272 & 0.3403 & 5.2679 & 8.3788 \\
LightGBM & 0.8723 & 0.3507 & 5.5308 & 8.7145 \\
PatchTST & \textbf{0.8086} & 0.3381 & \textbf{5.1435} & \textbf{8.0364} \\
\midrule
\multicolumn{5}{l}{\textbf{Statistical Baselines}} \\
AutoETS & 0.9066 & 0.4005 & 5.8721 & 8.9064 \\
Seasonal Naive & 1.0750 & 0.4934 & 6.9671 & 10.6799 \\
\bottomrule
\end{tabular}
\end{small}
\end{table}
\begin{table}[H]
\caption{PM2.5 leaderboard --- AURN}
\centering
\begin{small}
\begin{tabular}{lrrrr}
\toprule
model & MASE & CRPS & MAE & RMSE \\
\midrule
\multicolumn{5}{l}{\textbf{TSFMs}} \\
Chronos-2 & 0.8385 & 0.3696 & 3.1806 & 5.1566 \\
Chronos-Bolt & 0.8523 & 0.3772 & 3.2286 & 5.2664 \\
Kairos & 0.9130 & 0.4201 & 3.5345 & 5.5554 \\
Moirai-1 & 0.8438 & 0.3789 & 3.2118 & 5.2348 \\
Moirai-2 & 0.8253 & \underline{0.3582} & 3.1172 & 5.0622 \\
Sundial & 0.8328 & 0.4042 & 3.1815 & 5.0857 \\
TiRex & \textit{0.8196} & 0.3618 & \textit{3.1056} & 5.0375 \\
TimesFM-1.0 & 0.8627 & 0.3914 & 3.2936 & 5.2673 \\
TimesFM-2.0 & 0.8302 & 0.3710 & 3.1588 & 5.0958 \\
TimesFM-2.5 & \textbf{0.8172} & \textbf{0.3579} & \textbf{3.0898} & \textit{5.0191} \\
VisionTS++ & \underline{0.8180} & \textit{0.3615} & \underline{3.0913} & \underline{5.0151} \\
\midrule
\multicolumn{5}{l}{\textbf{ML Baselines}} \\
DLinear & 0.8542 & 0.4022 & 3.3123 & 5.1084 \\
DeepAR & 0.8490 & 0.3845 & 3.2414 & 5.2341 \\
LightGBM & 0.9087 & 0.3829 & 3.4312 & 5.6275 \\
PatchTST & 0.8258 & 0.3630 & 3.1088 & \textbf{4.9867} \\
\midrule
\multicolumn{5}{l}{\textbf{Statistical Baselines}} \\
AutoETS & 0.9254 & 0.4397 & 3.5609 & 5.5985 \\
Seasonal Naive & 1.1003 & 0.5557 & 4.2612 & 6.7161 \\
\bottomrule
\end{tabular}
\end{small}
\end{table}
\begin{table}[H]
\caption{SO2 leaderboard --- AURN}
\centering
\begin{small}
\begin{tabular}{lrrrr}
\toprule
model & MASE & CRPS & MAE & RMSE \\
\midrule
\multicolumn{5}{l}{\textbf{TSFMs}} \\
Chronos-2 & 1.0549 & 0.5089 & 0.7314 & 3.7773 \\
Chronos-Bolt & 1.0399 & 0.4979 & 0.7272 & 3.7711 \\
Kairos & 1.0768 & 0.5594 & 0.7608 & 3.7498 \\
Moirai-1 & 1.0407 & 0.4764 & 0.7090 & 3.7182 \\
Moirai-2 & 1.0371 & 0.4640 & 0.7003 & 3.7008 \\
Sundial & 1.0253 & 0.5818 & 0.7445 & \textbf{3.6520} \\
TiRex & \textit{1.0201} & 0.4723 & 0.6963 & 3.7210 \\
TimesFM-1.0 & 1.0323 & 0.4692 & 0.6985 & \textit{3.6541} \\
TimesFM-2.0 & 1.0461 & 0.4733 & \textit{0.6948} & \underline{3.6538} \\
TimesFM-2.5 & 1.0289 & \textit{0.4592} & 0.6977 & 3.6845 \\
VisionTS++ & 1.0268 & 0.4796 & 0.7056 & 3.7106 \\
\midrule
\multicolumn{5}{l}{\textbf{ML Baselines}} \\
DLinear & 1.0607 & 0.5956 & 0.8311 & 3.6892 \\
DeepAR & \textbf{1.0071} & \textbf{0.4358} & \textbf{0.6777} & 3.6793 \\
LightGBM & 1.0310 & 0.4958 & 0.7098 & 3.6604 \\
PatchTST & \underline{1.0187} & \underline{0.4438} & \underline{0.6836} & 3.6699 \\
\midrule
\multicolumn{5}{l}{\textbf{Statistical Baselines}} \\
AutoETS & 1.2048 & 1.0229 & 1.0786 & 4.2029 \\
Seasonal Naive & 1.3008 & 1.2936 & 1.0486 & 4.9559 \\
\bottomrule
\end{tabular}
\end{small}
\end{table}

\FloatBarrier
\subsection{CNEMC}
\begin{table}[H]
\caption{CO leaderboard --- CNEMC}
\centering
\begin{small}
\begin{tabular}{lrrrr}
\toprule
model & MASE & CRPS & MAE & RMSE \\
\midrule
\multicolumn{5}{l}{\textbf{TSFMs}} \\
Chronos-2 & 0.8536 & 0.1781 & 0.1379 & 0.2191 \\
Chronos-Bolt & 0.8583 & 0.1796 & 0.1386 & 0.2202 \\
Kairos & 1.0337 & 0.2194 & 0.1680 & 0.2504 \\
Moirai-1 & 0.8642 & 0.1821 & 0.1391 & 0.2192 \\
Moirai-2 & 0.8458 & \textit{0.1759} & 0.1360 & 0.2139 \\
Sundial & 0.8585 & 0.1936 & 0.1386 & 0.2138 \\
TiRex & \textit{0.8432} & 0.1771 & 0.1364 & 0.2162 \\
TimesFM-1.0 & 0.8905 & 0.1879 & 0.1433 & 0.2217 \\
TimesFM-2.0 & 0.8664 & 0.1812 & 0.1392 & 0.2181 \\
TimesFM-2.5 & \underline{0.8416} & \textbf{0.1751} & \underline{0.1351} & \textit{0.2129} \\
VisionTS++ & \textbf{0.8351} & \underline{0.1752} & \textbf{0.1344} & \underline{0.2118} \\
\midrule
\multicolumn{5}{l}{\textbf{ML Baselines}} \\
DLinear & 0.8698 & 0.1865 & 0.1398 & 0.2131 \\
DeepAR & 0.9161 & 0.1879 & 0.1473 & 0.2350 \\
LightGBM & 0.9122 & 0.1893 & 0.1455 & 0.2238 \\
PatchTST & 0.8472 & 0.1762 & \textit{0.1356} & \textbf{0.2109} \\
\midrule
\multicolumn{5}{l}{\textbf{Statistical Baselines}} \\
AutoETS & 0.9559 & 0.2100 & 0.1563 & 0.2331 \\
Seasonal Naive & 1.0727 & 0.2553 & 0.1795 & 0.2805 \\
\bottomrule
\end{tabular}
\end{small}
\end{table}
\begin{table}[H]
\caption{NO2 leaderboard --- CNEMC}
\centering
\begin{small}
\begin{tabular}{lrrrr}
\toprule
model & MASE & CRPS & MAE & RMSE \\
\midrule
\multicolumn{5}{l}{\textbf{TSFMs}} \\
Chronos-2 & 0.8188 & 0.2914 & 7.4805 & 11.3595 \\
Chronos-Bolt & 0.8208 & 0.2952 & 7.5027 & 11.4165 \\
Kairos & 1.1380 & 0.4147 & 10.4930 & 14.7121 \\
Moirai-1 & 0.8287 & 0.2990 & 7.5771 & 11.3383 \\
Moirai-2 & \textit{0.8082} & \textit{0.2871} & \textit{7.3716} & 11.1477 \\
Sundial & 0.8332 & 0.3269 & 7.6515 & 11.3152 \\
TiRex & 0.8204 & 0.2931 & 7.5177 & 11.3714 \\
TimesFM-1.0 & 0.8432 & 0.3069 & 7.7486 & 11.5995 \\
TimesFM-2.0 & 0.8226 & 0.2936 & 7.5105 & 11.3686 \\
TimesFM-2.5 & 0.8098 & 0.2873 & 7.3840 & 11.1352 \\
VisionTS++ & \textbf{0.8003} & \underline{0.2856} & \textbf{7.3098} & \textit{11.0871} \\
\midrule
\multicolumn{5}{l}{\textbf{ML Baselines}} \\
DLinear & 0.8184 & 0.2978 & 7.4954 & \underline{11.0636} \\
DeepAR & 0.8315 & 0.2942 & 7.5725 & 11.3417 \\
LightGBM & 0.8548 & 0.3145 & 7.8826 & 11.5634 \\
PatchTST & \underline{0.8035} & \textbf{0.2837} & \underline{7.3271} & \textbf{10.9889} \\
\midrule
\multicolumn{5}{l}{\textbf{Statistical Baselines}} \\
AutoETS & 0.9359 & 0.3595 & 8.6695 & 12.3952 \\
Seasonal Naive & 1.0563 & 0.4285 & 9.8504 & 14.7252 \\
\bottomrule
\end{tabular}
\end{small}
\end{table}
\begin{table}[H]
\caption{Ozone leaderboard --- CNEMC}
\centering
\begin{small}
\begin{tabular}{lrrrr}
\toprule
model & MASE & CRPS & MAE & RMSE \\
\midrule
\multicolumn{5}{l}{\textbf{TSFMs}} \\
Chronos-2 & 0.8515 & 0.2380 & 17.5516 & 24.4027 \\
Chronos-Bolt & 0.8465 & \textit{0.2372} & 17.4513 & 24.2806 \\
Kairos & 1.4097 & 0.3832 & 28.7210 & 38.4039 \\
Moirai-1 & 0.8823 & 0.2524 & 18.2204 & 25.0719 \\
Moirai-2 & 0.8648 & 0.2427 & 17.8240 & 24.6095 \\
Sundial & 0.8527 & 0.2600 & 17.5337 & \textit{23.8977} \\
TiRex & \underline{0.8412} & \underline{0.2362} & \textit{17.3516} & 24.1559 \\
TimesFM-1.0 & 0.9110 & 0.2586 & 18.7911 & 25.8213 \\
TimesFM-2.0 & 0.8620 & 0.2454 & 17.7385 & 24.3700 \\
TimesFM-2.5 & \textit{0.8428} & 0.2377 & \underline{17.3373} & \underline{23.8830} \\
VisionTS++ & 0.8445 & \textbf{0.2361} & 17.4011 & 24.0907 \\
\midrule
\multicolumn{5}{l}{\textbf{ML Baselines}} \\
DLinear & 0.8904 & 0.2466 & 18.1854 & 24.5150 \\
DeepAR & 0.8872 & 0.2667 & 18.2978 & 24.6612 \\
LightGBM & 0.8760 & 0.2448 & 17.9704 & 24.1388 \\
PatchTST & \textbf{0.8303} & 0.2388 & \textbf{17.0237} & \textbf{23.0460} \\
\midrule
\multicolumn{5}{l}{\textbf{Statistical Baselines}} \\
AutoETS & 0.9649 & 0.2798 & 19.9135 & 26.3167 \\
Seasonal Naive & 1.0495 & 0.3095 & 21.8108 & 29.5152 \\
\bottomrule
\end{tabular}
\end{small}
\end{table}
\begin{table}[H]
\caption{PM10 leaderboard --- CNEMC}
\centering
\begin{small}
\begin{tabular}{lrrrr}
\toprule
model & MASE & CRPS & MAE & RMSE \\
\midrule
\multicolumn{5}{l}{\textbf{TSFMs}} \\
Chronos-2 & 0.8576 & 0.2957 & 21.9493 & 51.5332 \\
Chronos-Bolt & 0.8746 & 0.3041 & 22.7016 & 55.8858 \\
Kairos & 0.9850 & 0.3582 & 25.1144 & 52.3456 \\
Moirai-1 & 0.8616 & 0.3031 & 22.0436 & 49.5990 \\
Moirai-2 & 0.8449 & \underline{0.2886} & 21.3591 & \textit{48.5309} \\
Sundial & 0.8623 & 0.3296 & 22.1074 & 48.7871 \\
TiRex & 0.8397 & 0.2921 & \textit{21.2632} & \underline{48.2206} \\
TimesFM-1.0 & 0.8755 & 0.3102 & 22.3289 & 49.3871 \\
TimesFM-2.0 & 0.8575 & 0.3018 & 21.8109 & 48.6662 \\
TimesFM-2.5 & \textit{0.8380} & \textit{0.2887} & \underline{21.1425} & \textbf{48.0782} \\
VisionTS++ & \underline{0.8353} & 0.2897 & 21.2954 & 48.6429 \\
\midrule
\multicolumn{5}{l}{\textbf{ML Baselines}} \\
DLinear & 0.8580 & 0.3019 & 22.3329 & 49.1409 \\
DeepAR & 0.8828 & 0.3090 & 22.6221 & 50.0531 \\
LightGBM & 0.9096 & 0.3166 & 23.4496 & 51.0299 \\
PatchTST & \textbf{0.8335} & \textbf{0.2809} & \textbf{21.1055} & 48.6913 \\
\midrule
\multicolumn{5}{l}{\textbf{Statistical Baselines}} \\
AutoETS & 0.9604 & 0.3771 & 25.9448 & 54.8411 \\
Seasonal Naive & 1.1213 & 0.4713 & 30.2852 & 65.9397 \\
\bottomrule
\end{tabular}
\end{small}
\end{table}
\begin{table}[H]
\caption{PM2.5 leaderboard --- CNEMC}
\centering
\begin{small}
\begin{tabular}{lrrrr}
\toprule
model & MASE & CRPS & MAE & RMSE \\
\midrule
\multicolumn{5}{l}{\textbf{TSFMs}} \\
Chronos-2 & 0.8343 & 0.3333 & 11.5998 & 23.2657 \\
Chronos-Bolt & 0.8493 & 0.3429 & 11.8942 & 24.5609 \\
Kairos & 0.9669 & 0.4160 & 13.5590 & 23.5243 \\
Moirai-1 & 0.8385 & 0.3449 & 11.6368 & 21.7274 \\
Moirai-2 & 0.8240 & \textbf{0.3278} & 11.3236 & \underline{21.0828} \\
Sundial & 0.8402 & 0.3766 & 11.6636 & 21.2644 \\
TiRex & \textit{0.8190} & 0.3316 & \textit{11.3072} & 21.1466 \\
TimesFM-1.0 & 0.8561 & 0.3567 & 11.8994 & 21.7923 \\
TimesFM-2.0 & 0.8341 & 0.3424 & 11.5130 & 21.2348 \\
TimesFM-2.5 & \underline{0.8151} & \underline{0.3284} & \textbf{11.1904} & \textbf{20.8240} \\
VisionTS++ & \textbf{0.8137} & \textit{0.3295} & \underline{11.2294} & \textit{21.1204} \\
\midrule
\multicolumn{5}{l}{\textbf{ML Baselines}} \\
DLinear & 0.8438 & 0.3560 & 11.8049 & 21.2023 \\
DeepAR & 0.8467 & 0.3402 & 11.6795 & 21.7003 \\
LightGBM & 0.9128 & 0.3613 & 12.6479 & 23.0049 \\
PatchTST & 0.8337 & 0.3356 & 11.5250 & 21.3781 \\
\midrule
\multicolumn{5}{l}{\textbf{Statistical Baselines}} \\
AutoETS & 0.9317 & 0.4187 & 13.3167 & 24.0191 \\
Seasonal Naive & 1.0977 & 0.5284 & 15.8193 & 28.9843 \\
\bottomrule
\end{tabular}
\end{small}
\end{table}
\begin{table}[H]
\caption{SO2 leaderboard --- CNEMC}
\centering
\begin{small}
\begin{tabular}{lrrrr}
\toprule
model & MASE & CRPS & MAE & RMSE \\
\midrule
\multicolumn{5}{l}{\textbf{TSFMs}} \\
Chronos-2 & 0.9330 & 0.2042 & 2.1744 & 4.5784 \\
Chronos-Bolt & 0.9395 & 0.2059 & 2.1863 & 4.6160 \\
Kairos & 1.0599 & 0.2352 & 2.5095 & 4.9840 \\
Moirai-1 & 0.9437 & 0.2064 & 2.1916 & 4.6164 \\
Moirai-2 & 0.9295 & 0.2013 & 2.1559 & 4.5579 \\
Sundial & 0.9452 & 0.2263 & 2.2157 & \underline{4.5111} \\
TiRex & \underline{0.9230} & \underline{0.2007} & \textit{2.1433} & 4.5606 \\
TimesFM-1.0 & 0.9740 & 0.2113 & 2.2277 & 4.6152 \\
TimesFM-2.0 & 0.9470 & 0.2037 & 2.1788 & 4.6132 \\
TimesFM-2.5 & \textit{0.9262} & \textbf{0.1991} & \underline{2.1410} & 4.5618 \\
VisionTS++ & \textbf{0.9204} & \textit{0.2009} & \textbf{2.1295} & \textbf{4.5077} \\
\midrule
\multicolumn{5}{l}{\textbf{ML Baselines}} \\
DLinear & 1.1207 & 0.2282 & 2.3040 & \textit{4.5377} \\
DeepAR & 1.1635 & 0.2082 & 2.2860 & 4.8613 \\
LightGBM & 1.1722 & 0.2161 & 2.2416 & 4.5945 \\
PatchTST & 1.0139 & 0.2033 & 2.1704 & 4.5816 \\
\midrule
\multicolumn{5}{l}{\textbf{Statistical Baselines}} \\
AutoETS & 1.0628 & 0.2658 & 2.5879 & 4.8624 \\
Seasonal Naive & 1.1584 & 0.3282 & 2.8828 & 5.9823 \\
\bottomrule
\end{tabular}
\end{small}
\end{table}

\FloatBarrier
\subsection{CPCB}
\begin{table}[H]
\caption{CO leaderboard --- CPCB}
\centering
\begin{small}
\begin{tabular}{lrrrr}
\toprule
model & MASE & CRPS & MAE & RMSE \\
\midrule
\multicolumn{5}{l}{\textbf{TSFMs}} \\
Chronos-2 & 0.9244 & 0.2618 & 0.2482 & 0.4584 \\
Chronos-Bolt & 0.9419 & 0.2700 & 0.2515 & 0.4678 \\
Kairos & 1.2723 & 0.3652 & 0.3325 & 0.5519 \\
Moirai-1 & 0.9621 & 0.2810 & 0.2547 & 0.4565 \\
Moirai-2 & 0.9172 & \textbf{0.2555} & 0.2452 & 0.4431 \\
Sundial & 0.9103 & 0.2898 & \textit{0.2428} & \textbf{0.4342} \\
TiRex & \underline{0.9060} & \textit{0.2586} & 0.2434 & 0.4453 \\
TimesFM-1.0 & 0.9476 & 0.2844 & 0.2512 & 0.4502 \\
TimesFM-2.0 & 0.9228 & 0.2716 & 0.2454 & 0.4454 \\
TimesFM-2.5 & \textbf{0.9043} & 0.2612 & \textbf{0.2414} & \underline{0.4380} \\
VisionTS++ & \textit{0.9074} & \underline{0.2572} & \underline{0.2421} & 0.4419 \\
\midrule
\multicolumn{5}{l}{\textbf{ML Baselines}} \\
DLinear & 1.0502 & 0.3095 & 0.2581 & 0.4456 \\
DeepAR & 0.9918 & 0.2587 & 0.2507 & 0.4475 \\
LightGBM & 0.9945 & 0.3036 & 0.2534 & 0.4428 \\
PatchTST & 0.9779 & 0.2683 & 0.2476 & \textit{0.4413} \\
\midrule
\multicolumn{5}{l}{\textbf{Statistical Baselines}} \\
AutoETS & 1.0189 & 0.3330 & 0.2751 & 0.4686 \\
Seasonal Naive & 1.1246 & 0.4027 & 0.3069 & 0.5545 \\
\bottomrule
\end{tabular}
\end{small}
\end{table}
\begin{table}[H]
\caption{NO2 leaderboard --- CPCB}
\centering
\begin{small}
\begin{tabular}{lrrrr}
\toprule
model & MASE & CRPS & MAE & RMSE \\
\midrule
\multicolumn{5}{l}{\textbf{TSFMs}} \\
Chronos-2 & 1.0906 & 0.1619 & 4.9475 & 10.2941 \\
Chronos-Bolt & 1.0990 & 0.1685 & 4.9901 & 10.4766 \\
Kairos & 1.4645 & 0.2509 & 7.4305 & 13.2823 \\
Moirai-1 & 1.1209 & 0.1756 & 5.1044 & 10.3942 \\
Moirai-2 & 1.0900 & \textit{0.1616} & \textit{4.9067} & \underline{10.0536} \\
Sundial & 1.1141 & 0.1922 & 5.0875 & 10.1572 \\
TiRex & \underline{1.0854} & 0.1638 & 4.9244 & 10.1495 \\
TimesFM-1.0 & 1.1952 & 0.1763 & 5.1509 & 10.4106 \\
TimesFM-2.0 & 1.1318 & 0.1699 & 5.0297 & 10.2901 \\
TimesFM-2.5 & \textit{1.0865} & \textbf{0.1593} & \underline{4.8959} & \textbf{10.0528} \\
VisionTS++ & \textbf{1.0806} & \underline{0.1609} & \textbf{4.8543} & \textit{10.0593} \\
\midrule
\multicolumn{5}{l}{\textbf{ML Baselines}} \\
DLinear & 1.7975 & 0.2402 & 5.2552 & 10.2494 \\
DeepAR & 2.7737 & 0.2075 & 5.5075 & 10.6324 \\
LightGBM & 1.6227 & 0.2091 & 5.2668 & 10.5148 \\
PatchTST & 1.5473 & 0.1839 & 5.2061 & 10.5882 \\
\midrule
\multicolumn{5}{l}{\textbf{Statistical Baselines}} \\
AutoETS & 1.2158 & 0.3116 & 5.7697 & 10.9152 \\
Seasonal Naive & 1.3420 & 0.3745 & 6.3035 & 12.8719 \\
\bottomrule
\end{tabular}
\end{small}
\end{table}
\begin{table}[H]
\caption{Ozone leaderboard --- CPCB}
\centering
\begin{small}
\begin{tabular}{lrrrr}
\toprule
model & MASE & CRPS & MAE & RMSE \\
\midrule
\multicolumn{5}{l}{\textbf{TSFMs}} \\
Chronos-2 & \underline{0.9423} & \textit{0.2266} & \textit{7.4354} & 13.7858 \\
Chronos-Bolt & 0.9739 & 0.2337 & 7.5221 & 13.8563 \\
Kairos & 2.6636 & 0.4313 & 14.7043 & 23.7030 \\
Moirai-1 & 1.0567 & 0.2422 & 7.8403 & 14.2222 \\
Moirai-2 & 0.9789 & 0.2272 & 7.5333 & 13.8471 \\
Sundial & 0.9909 & 0.2591 & 7.6890 & 13.7808 \\
TiRex & \textbf{0.9391} & \underline{0.2263} & \underline{7.3793} & \underline{13.6980} \\
TimesFM-1.0 & 1.1591 & 0.2536 & 8.1274 & 14.5997 \\
TimesFM-2.0 & 1.0246 & 0.2426 & 7.7659 & 14.1007 \\
TimesFM-2.5 & 0.9578 & 0.2279 & 7.4962 & \textit{13.7385} \\
VisionTS++ & \textit{0.9534} & \textbf{0.2240} & \textbf{7.3577} & \textbf{13.6291} \\
\midrule
\multicolumn{5}{l}{\textbf{ML Baselines}} \\
DLinear & 1.5990 & 0.2816 & 7.9252 & 13.8600 \\
DeepAR & 2.1098 & 0.2424 & 7.9193 & 14.1435 \\
LightGBM & 1.0854 & 0.2533 & 7.7494 & 14.0969 \\
PatchTST & 1.2962 & 0.2388 & 7.7170 & 13.8499 \\
\midrule
\multicolumn{5}{l}{\textbf{Statistical Baselines}} \\
AutoETS & 1.0750 & 0.3011 & 8.6509 & 14.6753 \\
Seasonal Naive & 1.1367 & 0.3663 & 9.1527 & 16.8007 \\
\bottomrule
\end{tabular}
\end{small}
\end{table}
\begin{table}[H]
\caption{PM10 leaderboard --- CPCB}
\centering
\begin{small}
\begin{tabular}{lrrrr}
\toprule
model & MASE & CRPS & MAE & RMSE \\
\midrule
\multicolumn{5}{l}{\textbf{TSFMs}} \\
Chronos-2 & 0.8702 & 0.2302 & 32.4988 & 54.4752 \\
Chronos-Bolt & 0.8785 & 0.2344 & 32.7621 & 55.6797 \\
Kairos & 1.1371 & 0.3002 & 42.8265 & 65.5376 \\
Moirai-1 & 0.8913 & 0.2379 & 33.0893 & 54.1614 \\
Moirai-2 & 0.8636 & \textit{0.2284} & 32.1472 & 53.0072 \\
Sundial & 0.8816 & 0.2566 & 32.9812 & 53.3992 \\
TiRex & \textit{0.8632} & 0.2312 & 32.2666 & 53.4504 \\
TimesFM-1.0 & 0.9162 & 0.2414 & 33.4620 & 54.7526 \\
TimesFM-2.0 & 0.8756 & 0.2335 & 32.5497 & 53.7138 \\
TimesFM-2.5 & \underline{0.8600} & \underline{0.2276} & \underline{32.0420} & \underline{52.9477} \\
VisionTS++ & \textbf{0.8495} & \textbf{0.2263} & \textbf{31.6192} & \textbf{52.5272} \\
\midrule
\multicolumn{5}{l}{\textbf{ML Baselines}} \\
DLinear & 1.3259 & 0.2420 & 32.8112 & 52.9855 \\
DeepAR & 1.2195 & 0.2323 & 33.3015 & 55.2323 \\
LightGBM & 1.2375 & 0.2570 & 34.2894 & 54.6240 \\
PatchTST & 1.5061 & 0.2286 & \textit{32.1340} & \textit{52.9582} \\
\midrule
\multicolumn{5}{l}{\textbf{Statistical Baselines}} \\
AutoETS & 0.9688 & 0.2758 & 36.8008 & 57.6024 \\
Seasonal Naive & 1.0898 & 0.3289 & 41.5503 & 67.9091 \\
\bottomrule
\end{tabular}
\end{small}
\end{table}
\begin{table}[H]
\caption{PM2.5 leaderboard --- CPCB}
\centering
\begin{small}
\begin{tabular}{lrrrr}
\toprule
model & MASE & CRPS & MAE & RMSE \\
\midrule
\multicolumn{5}{l}{\textbf{TSFMs}} \\
Chronos-2 & 0.8530 & 0.2604 & 16.5335 & 32.1943 \\
Chronos-Bolt & 0.8643 & 0.3098 & 16.7812 & 34.4363 \\
Kairos & 1.1025 & 0.4086 & 22.2215 & 37.7633 \\
Moirai-1 & 0.8680 & 0.2845 & 16.7435 & 31.3103 \\
Moirai-2 & 0.8460 & \textit{0.2560} & \textit{16.2858} & \textit{30.6680} \\
Sundial & 0.8630 & 0.3287 & 16.6975 & 30.7959 \\
TiRex & \textit{0.8444} & 0.2658 & 16.3108 & 30.9252 \\
TimesFM-1.0 & 0.8780 & 0.3090 & 16.9908 & 31.6615 \\
TimesFM-2.0 & 0.8582 & 0.2815 & 16.4874 & 30.9618 \\
TimesFM-2.5 & \underline{0.8429} & \textbf{0.2550} & \underline{16.2342} & \underline{30.5685} \\
VisionTS++ & \textbf{0.8324} & \underline{0.2560} & \textbf{16.0400} & \textbf{30.4889} \\
\midrule
\multicolumn{5}{l}{\textbf{ML Baselines}} \\
DLinear & 0.9364 & 0.3477 & 16.8454 & 30.7986 \\
DeepAR & 0.9807 & 0.3353 & 16.9515 & 32.2515 \\
LightGBM & 0.9732 & 0.5845 & 17.8979 & 32.1285 \\
PatchTST & 0.9696 & 0.3268 & 16.8673 & 31.5649 \\
\midrule
\multicolumn{5}{l}{\textbf{Statistical Baselines}} \\
AutoETS & 0.9533 & 0.4163 & 18.9739 & 33.8174 \\
Seasonal Naive & 1.0807 & 0.5142 & 21.3757 & 40.2379 \\
\bottomrule
\end{tabular}
\end{small}
\end{table}
\begin{table}[H]
\caption{SO2 leaderboard --- CPCB}
\centering
\begin{small}
\begin{tabular}{lrrrr}
\toprule
model & MASE & CRPS & MAE & RMSE \\
\midrule
\multicolumn{5}{l}{\textbf{TSFMs}} \\
Chronos-2 & 1.1375 & 0.2115 & 3.1569 & 6.7563 \\
Chronos-Bolt & 1.1596 & 0.2149 & 3.2104 & 6.8318 \\
Kairos & 1.3247 & 0.2452 & 3.6668 & 7.1336 \\
Moirai-1 & 1.1886 & 0.2173 & 3.2560 & 6.6733 \\
Moirai-2 & 1.1344 & \underline{0.2055} & 3.1239 & 6.4565 \\
Sundial & 1.1305 & 0.2269 & 3.1224 & \textbf{6.3206} \\
TiRex & \textbf{1.1224} & 0.2065 & \underline{3.0898} & 6.4830 \\
TimesFM-1.0 & 1.1911 & 0.2188 & 3.2356 & 6.5849 \\
TimesFM-2.0 & 1.1516 & 0.2118 & 3.1552 & 6.5341 \\
TimesFM-2.5 & \underline{1.1250} & \textbf{0.2041} & \textbf{3.0852} & \underline{6.4062} \\
VisionTS++ & \textit{1.1302} & \textit{0.2065} & \textit{3.0977} & \textit{6.4233} \\
\midrule
\multicolumn{5}{l}{\textbf{ML Baselines}} \\
DLinear & 1.4425 & 0.2472 & 3.3913 & 6.5172 \\
DeepAR & 1.5580 & 0.2273 & 3.4437 & 6.7332 \\
LightGBM & 1.2072 & 0.2211 & 3.2386 & 6.5352 \\
PatchTST & 1.2174 & 0.2104 & 3.1911 & 6.5207 \\
\midrule
\multicolumn{5}{l}{\textbf{Statistical Baselines}} \\
AutoETS & 1.2292 & 0.2596 & 3.5339 & 6.8045 \\
Seasonal Naive & 1.3528 & 0.3217 & 3.9838 & 8.1470 \\
\bottomrule
\end{tabular}
\end{small}
\end{table}

\FloatBarrier
\subsection{EEA-Germany}
\begin{table}[H]
\caption{CO leaderboard --- EEA DE}
\centering
\begin{small}
\begin{tabular}{lrrrr}
\toprule
model & MASE & CRPS & MAE & RMSE \\
\midrule
\multicolumn{5}{l}{\textbf{TSFMs}} \\
Chronos-2 & 0.8426 & 0.1798 & 0.0644 & 0.1072 \\
Chronos-Bolt & 0.8393 & 0.1798 & 0.0639 & 0.1063 \\
Kairos & 1.0330 & 0.2181 & 0.0786 & 0.1207 \\
Moirai-1 & 0.8781 & 0.1875 & 0.0668 & 0.1091 \\
Moirai-2 & 0.8530 & 0.1808 & 0.0650 & 0.1068 \\
Sundial & 0.8436 & 0.1959 & 0.0645 & \textit{0.1049} \\
TiRex & \underline{0.8330} & \textit{0.1783} & \textit{0.0637} & 0.1057 \\
TimesFM-1.0 & 0.8580 & 0.1839 & 0.0654 & 0.1073 \\
TimesFM-2.0 & 0.8437 & 0.1795 & 0.0642 & 0.1064 \\
TimesFM-2.5 & \textit{0.8371} & \underline{0.1771} & \underline{0.0636} & 0.1051 \\
VisionTS++ & \textbf{0.8252} & \textbf{0.1752} & \textbf{0.0628} & \textbf{0.1042} \\
\midrule
\multicolumn{5}{l}{\textbf{ML Baselines}} \\
DLinear & 0.8841 & 0.1935 & 0.0675 & 0.1053 \\
DeepAR & 0.9102 & 0.1878 & 0.0687 & 0.1133 \\
LightGBM & 0.9318 & 0.1920 & 0.0695 & 0.1137 \\
PatchTST & 0.8626 & 0.1832 & 0.0653 & \underline{0.1044} \\
\midrule
\multicolumn{5}{l}{\textbf{Statistical Baselines}} \\
AutoETS & 0.9533 & 0.2120 & 0.0736 & 0.1144 \\
Seasonal Naive & 1.0646 & 0.2475 & 0.0818 & 0.1324 \\
\bottomrule
\end{tabular}
\end{small}
\end{table}
\begin{table}[H]
\caption{NO2 leaderboard --- EEA DE}
\centering
\begin{small}
\begin{tabular}{lrrrr}
\toprule
model & MASE & CRPS & MAE & RMSE \\
\midrule
\multicolumn{5}{l}{\textbf{TSFMs}} \\
Chronos-2 & 0.8219 & 0.3220 & 5.3819 & 7.8577 \\
Chronos-Bolt & \underline{0.8168} & 0.3229 & \underline{5.3448} & 7.7813 \\
Kairos & 1.0379 & 0.4088 & 6.8216 & 9.3895 \\
Moirai-1 & 0.8443 & 0.3331 & 5.5615 & 8.0232 \\
Moirai-2 & 0.8302 & 0.3215 & 5.4656 & 7.9456 \\
Sundial & 0.8284 & 0.3592 & 5.4487 & 7.8318 \\
TiRex & 0.8227 & 0.3237 & 5.4129 & 7.8947 \\
TimesFM-1.0 & 0.8350 & 0.3331 & 5.4824 & 7.9628 \\
TimesFM-2.0 & 0.8208 & 0.3209 & 5.3786 & 7.8830 \\
TimesFM-2.5 & \textit{0.8171} & \underline{0.3171} & \textit{5.3644} & 7.8196 \\
VisionTS++ & \textbf{0.8082} & \textbf{0.3164} & \textbf{5.2999} & \textit{7.7330} \\
\midrule
\multicolumn{5}{l}{\textbf{ML Baselines}} \\
DLinear & 0.8401 & 0.3388 & 5.5012 & \textbf{7.6941} \\
DeepAR & 0.8583 & \textit{0.3183} & 5.5993 & 8.3161 \\
LightGBM & 0.8825 & 0.3425 & 5.7595 & 8.3442 \\
PatchTST & 0.8410 & 0.3359 & 5.5018 & \underline{7.7037} \\
\midrule
\multicolumn{5}{l}{\textbf{Statistical Baselines}} \\
AutoETS & 0.9284 & 0.3863 & 6.1033 & 8.4334 \\
Seasonal Naive & 1.0613 & 0.4610 & 6.9819 & 9.8980 \\
\bottomrule
\end{tabular}
\end{small}
\end{table}
\begin{table}[H]
\caption{Ozone leaderboard --- EEA DE}
\centering
\begin{small}
\begin{tabular}{lrrrr}
\toprule
model & MASE & CRPS & MAE & RMSE \\
\midrule
\multicolumn{5}{l}{\textbf{TSFMs}} \\
Chronos-2 & 0.8090 & \textit{0.2326} & 12.5185 & 16.8718 \\
Chronos-Bolt & 0.8078 & 0.2346 & 12.4934 & 16.8510 \\
Kairos & 1.2103 & 0.3442 & 18.9412 & 25.1775 \\
Moirai-1 & 0.8362 & 0.2440 & 12.9755 & 17.2611 \\
Moirai-2 & 0.8272 & 0.2440 & 12.8443 & 17.2177 \\
Sundial & 0.8108 & 0.2651 & 12.5311 & \textit{16.5115} \\
TiRex & \textbf{0.7898} & \textbf{0.2318} & \textbf{12.2285} & \underline{16.4471} \\
TimesFM-1.0 & 0.8606 & 0.2615 & 13.3800 & 17.7111 \\
TimesFM-2.0 & 0.8090 & 0.2412 & 12.5215 & 16.6099 \\
TimesFM-2.5 & \underline{0.7949} & 0.2336 & \underline{12.3024} & \textbf{16.3958} \\
VisionTS++ & \textit{0.8005} & \underline{0.2323} & \textit{12.3883} & 16.5657 \\
\midrule
\multicolumn{5}{l}{\textbf{ML Baselines}} \\
DLinear & 0.8216 & 0.2504 & 12.6760 & 16.5773 \\
DeepAR & 0.8385 & 0.2711 & 12.9442 & 16.9954 \\
LightGBM & 1.0149 & 0.3021 & 15.5326 & 19.3793 \\
PatchTST & 0.8158 & 0.2610 & 12.6040 & 16.5439 \\
\midrule
\multicolumn{5}{l}{\textbf{Statistical Baselines}} \\
AutoETS & 0.9091 & 0.2740 & 14.0778 & 18.1145 \\
Seasonal Naive & 1.0482 & 0.3248 & 16.3203 & 21.3005 \\
\bottomrule
\end{tabular}
\end{small}
\end{table}
\begin{table}[H]
\caption{PM10 leaderboard --- EEA DE}
\centering
\begin{small}
\begin{tabular}{lrrrr}
\toprule
model & MASE & CRPS & MAE & RMSE \\
\midrule
\multicolumn{5}{l}{\textbf{TSFMs}} \\
Chronos-2 & 0.8343 & 0.3130 & 5.0470 & 9.1636 \\
Chronos-Bolt & 0.8445 & 0.3183 & 5.1055 & 9.2909 \\
Kairos & 0.9354 & 0.3616 & 5.6878 & 9.5998 \\
Moirai-1 & 0.8397 & 0.3182 & 5.0686 & 9.0318 \\
Moirai-2 & 0.8287 & \textit{0.3048} & 4.9681 & 8.8505 \\
Sundial & 0.8407 & 0.3462 & 5.0761 & 8.8550 \\
TiRex & 0.8246 & 0.3084 & 4.9517 & 8.8209 \\
TimesFM-1.0 & 0.8601 & 0.3302 & 5.1837 & 9.0384 \\
TimesFM-2.0 & 0.8372 & 0.3166 & 5.0500 & 8.9691 \\
TimesFM-2.5 & \textit{0.8187} & \underline{0.3021} & \underline{4.8999} & \underline{8.7279} \\
VisionTS++ & \underline{0.8177} & 0.3064 & \textit{4.9071} & \textit{8.7896} \\
\midrule
\multicolumn{5}{l}{\textbf{ML Baselines}} \\
DLinear & 0.8303 & 0.3163 & 5.0137 & \textbf{8.7273} \\
DeepAR & 0.8602 & 0.3305 & 5.2019 & 9.1086 \\
LightGBM & 0.9093 & 0.3527 & 5.5419 & 9.4463 \\
PatchTST & \textbf{0.8155} & \textbf{0.3010} & \textbf{4.8939} & 8.8157 \\
\midrule
\multicolumn{5}{l}{\textbf{Statistical Baselines}} \\
AutoETS & 0.9243 & 0.3816 & 5.6784 & 9.6087 \\
Seasonal Naive & 1.0899 & 0.4780 & 6.7188 & 11.5799 \\
\bottomrule
\end{tabular}
\end{small}
\end{table}
\begin{table}[H]
\caption{PM2.5 leaderboard --- EEA DE}
\centering
\begin{small}
\begin{tabular}{lrrrr}
\toprule
model & MASE & CRPS & MAE & RMSE \\
\midrule
\multicolumn{5}{l}{\textbf{TSFMs}} \\
Chronos-2 & 0.8324 & 0.3375 & 3.3436 & 5.9400 \\
Chronos-Bolt & 0.8476 & 0.3465 & 3.4119 & 6.0666 \\
Kairos & 0.9508 & 0.4088 & 3.8665 & 6.3014 \\
Moirai-1 & 0.8391 & 0.3489 & 3.3854 & 5.8574 \\
Moirai-2 & 0.8243 & \textbf{0.3304} & 3.2947 & 5.6964 \\
Sundial & 0.8448 & 0.3826 & 3.4047 & 5.7638 \\
TiRex & \textit{0.8206} & \textit{0.3337} & \textit{3.2833} & 5.6727 \\
TimesFM-1.0 & 0.8688 & 0.3679 & 3.5036 & 5.9506 \\
TimesFM-2.0 & 0.8330 & 0.3456 & 3.3427 & 5.7231 \\
TimesFM-2.5 & \textbf{0.8163} & \underline{0.3307} & \textbf{3.2621} & \underline{5.6238} \\
VisionTS++ & \underline{0.8179} & 0.3344 & \underline{3.2699} & \textit{5.6720} \\
\midrule
\multicolumn{5}{l}{\textbf{ML Baselines}} \\
DLinear & 0.8252 & 0.3424 & 3.3132 & \textbf{5.6214} \\
DeepAR & 0.8979 & 0.3747 & 3.6470 & 6.2293 \\
LightGBM & 0.9464 & 0.4412 & 3.9312 & 6.3485 \\
PatchTST & 0.8463 & 0.3375 & 3.3757 & 5.7457 \\
\midrule
\multicolumn{5}{l}{\textbf{Statistical Baselines}} \\
AutoETS & 0.9290 & 0.4193 & 3.7812 & 6.2433 \\
Seasonal Naive & 1.1014 & 0.5335 & 4.5330 & 7.6529 \\
\bottomrule
\end{tabular}
\end{small}
\end{table}
\begin{table}[H]
\caption{SO2 leaderboard --- EEA DE}
\centering
\begin{small}
\begin{tabular}{lrrrr}
\toprule
model & MASE & CRPS & MAE & RMSE \\
\midrule
\multicolumn{5}{l}{\textbf{TSFMs}} \\
Chronos-2 & 1.1408 & 0.3855 & 0.8246 & 2.5037 \\
Chronos-Bolt & 1.1519 & 0.3872 & 0.8230 & 2.5826 \\
Kairos & 1.2037 & 0.5267 & 0.8707 & 2.5357 \\
Moirai-1 & 1.1489 & 0.4005 & 0.8127 & 2.5060 \\
Moirai-2 & \textit{1.1236} & \underline{0.3640} & \textit{0.8014} & 2.4900 \\
Sundial & 1.1461 & 0.4773 & 0.8518 & \textbf{2.4232} \\
TiRex & \textbf{1.1158} & \textit{0.3690} & \textbf{0.7872} & \textit{2.4648} \\
TimesFM-1.0 & 1.4292 & 0.3899 & 0.8163 & 2.4760 \\
TimesFM-2.0 & 1.2457 & 0.4151 & 0.8065 & 2.4744 \\
TimesFM-2.5 & 1.1255 & \textbf{0.3612} & \underline{0.7931} & 2.4648 \\
VisionTS++ & \underline{1.1184} & 0.3820 & 0.8090 & 2.4838 \\
\midrule
\multicolumn{5}{l}{\textbf{ML Baselines}} \\
DLinear & 3.1786 & 0.5793 & 0.9505 & \underline{2.4537} \\
DeepAR & 1.4626 & 0.4529 & 0.8335 & 2.5126 \\
LightGBM & 4.3892 & 0.4768 & 0.8417 & 2.4860 \\
PatchTST & 1.3691 & 0.4049 & 0.8104 & 2.4971 \\
\midrule
\multicolumn{5}{l}{\textbf{Statistical Baselines}} \\
AutoETS & 1.3345 & 0.7793 & 1.1364 & 2.6697 \\
Seasonal Naive & 1.4236 & 1.0430 & 1.1851 & 3.2927 \\
\bottomrule
\end{tabular}
\end{small}
\end{table}

\FloatBarrier
\subsection{EEA-France}
\begin{table}[H]
\caption{CO leaderboard --- EEA FR}
\centering
\begin{small}
\begin{tabular}{lrrrr}
\toprule
model & MASE & CRPS & MAE & RMSE \\
\midrule
\multicolumn{5}{l}{\textbf{TSFMs}} \\
Chronos-2 & 0.8230 & 0.1978 & 0.0571 & 0.0954 \\
Chronos-Bolt & 0.8259 & 0.1990 & 0.0571 & 0.0994 \\
Kairos & 0.9988 & 0.2401 & 0.0676 & 0.1044 \\
Moirai-1 & 0.8493 & 0.2062 & 0.0584 & 0.0955 \\
Moirai-2 & 0.8176 & \textit{0.1963} & 0.0563 & 0.0928 \\
Sundial & 0.8143 & 0.2127 & 0.0563 & \textit{0.0920} \\
TiRex & \textit{0.8058} & 0.1964 & 0.0561 & 0.0931 \\
TimesFM-1.0 & 0.8310 & 0.2039 & 0.0577 & 0.0950 \\
TimesFM-2.0 & 0.8162 & 0.1979 & \textit{0.0561} & 0.0928 \\
TimesFM-2.5 & \underline{0.8053} & \underline{0.1936} & \underline{0.0554} & 0.0921 \\
VisionTS++ & \textbf{0.7966} & \textbf{0.1917} & \textbf{0.0548} & \textbf{0.0908} \\
\midrule
\multicolumn{5}{l}{\textbf{ML Baselines}} \\
DLinear & 0.8224 & 0.2050 & 0.0567 & \underline{0.0909} \\
DeepAR & 0.8968 & 0.2189 & 0.0608 & 0.0954 \\
LightGBM & 0.9137 & 0.2236 & 0.0626 & 0.0966 \\
PatchTST & 0.8456 & 0.2062 & 0.0577 & 0.0924 \\
\midrule
\multicolumn{5}{l}{\textbf{Statistical Baselines}} \\
AutoETS & 0.9259 & 0.2296 & 0.0639 & 0.0997 \\
Seasonal Naive & 1.0679 & 0.2772 & 0.0745 & 0.1206 \\
\bottomrule
\end{tabular}
\end{small}
\end{table}
\begin{table}[H]
\caption{NO2 leaderboard --- EEA FR}
\centering
\begin{small}
\begin{tabular}{lrrrr}
\toprule
model & MASE & CRPS & MAE & RMSE \\
\midrule
\multicolumn{5}{l}{\textbf{TSFMs}} \\
Chronos-2 & 0.8275 & 0.3636 & 5.3837 & 8.2704 \\
Chronos-Bolt & \underline{0.8176} & 0.3629 & \underline{5.3119} & 8.1386 \\
Kairos & 1.1078 & 0.4820 & 7.2358 & 10.2930 \\
Moirai-1 & 0.8531 & 0.3777 & 5.5467 & 8.3977 \\
Moirai-2 & 0.8313 & 0.3633 & 5.4086 & 8.2653 \\
Sundial & 0.8350 & 0.4102 & 5.4487 & 8.2165 \\
TiRex & 0.8293 & 0.3667 & 5.4086 & 8.3011 \\
TimesFM-1.0 & 0.8384 & 0.3786 & 5.4695 & 8.3687 \\
TimesFM-2.0 & 0.8255 & 0.3599 & 5.3664 & 8.2945 \\
TimesFM-2.5 & \textit{0.8208} & \underline{0.3569} & \textit{5.3355} & 8.1879 \\
VisionTS++ & \textbf{0.8131} & \textbf{0.3560} & \textbf{5.2793} & \textit{8.1269} \\
\midrule
\multicolumn{5}{l}{\textbf{ML Baselines}} \\
DLinear & 0.8367 & 0.3884 & 5.4310 & \textbf{8.0152} \\
DeepAR & 0.8484 & \textit{0.3584} & 5.4845 & 8.4766 \\
LightGBM & 0.8789 & 0.4290 & 5.7480 & 8.3958 \\
PatchTST & 0.8357 & 0.3777 & 5.4206 & \underline{8.0296} \\
\midrule
\multicolumn{5}{l}{\textbf{Statistical Baselines}} \\
AutoETS & 0.9456 & 0.4554 & 6.1492 & 8.7922 \\
Seasonal Naive & 1.0544 & 0.5358 & 6.8872 & 10.2616 \\
\bottomrule
\end{tabular}
\end{small}
\end{table}
\begin{table}[H]
\caption{Ozone leaderboard --- EEA FR}
\centering
\begin{small}
\begin{tabular}{lrrrr}
\toprule
model & MASE & CRPS & MAE & RMSE \\
\midrule
\multicolumn{5}{l}{\textbf{TSFMs}} \\
Chronos-2 & 0.8146 & \textit{0.2062} & 12.0841 & 16.3168 \\
Chronos-Bolt & 0.8119 & 0.2064 & 12.0452 & 16.2661 \\
Kairos & 1.2111 & 0.3034 & 17.9576 & 23.5162 \\
Moirai-1 & 0.8450 & 0.2158 & 12.5607 & 16.6852 \\
Moirai-2 & 0.8332 & 0.2137 & 12.3981 & 16.6311 \\
Sundial & 0.8170 & 0.2298 & 12.1212 & 16.0334 \\
TiRex & \textbf{0.7951} & \textbf{0.2044} & \textbf{11.7964} & \underline{15.9004} \\
TimesFM-1.0 & 0.8733 & 0.2297 & 13.0399 & 17.2936 \\
TimesFM-2.0 & 0.8154 & 0.2123 & 12.0961 & 16.1078 \\
TimesFM-2.5 & \underline{0.8012} & 0.2066 & \underline{11.8760} & \textbf{15.8890} \\
VisionTS++ & \textit{0.8055} & \underline{0.2047} & \textit{11.9388} & \textit{16.0248} \\
\midrule
\multicolumn{5}{l}{\textbf{ML Baselines}} \\
DLinear & 0.8365 & 0.2299 & 12.3501 & 16.2921 \\
DeepAR & 1.1036 & 0.3080 & 16.0115 & 20.6528 \\
LightGBM & 0.8953 & 0.2360 & 13.2572 & 16.9690 \\
PatchTST & 0.8334 & 0.2227 & 12.3068 & 16.1720 \\
\midrule
\multicolumn{5}{l}{\textbf{Statistical Baselines}} \\
AutoETS & 0.9205 & 0.2409 & 13.6965 & 17.6835 \\
Seasonal Naive & 1.0497 & 0.2836 & 15.7446 & 20.7604 \\
\bottomrule
\end{tabular}
\end{small}
\end{table}
\begin{table}[H]
\caption{PM10 leaderboard --- EEA FR}
\centering
\begin{small}
\begin{tabular}{lrrrr}
\toprule
model & MASE & CRPS & MAE & RMSE \\
\midrule
\multicolumn{5}{l}{\textbf{TSFMs}} \\
Chronos-2 & 0.8385 & 0.3202 & 5.3807 & 8.9583 \\
Chronos-Bolt & 0.8409 & 0.3247 & 5.3937 & 9.0215 \\
Kairos & 0.9506 & 0.3699 & 6.1405 & 9.7315 \\
Moirai-1 & 0.8481 & 0.3257 & 5.4427 & 8.9578 \\
Moirai-2 & 0.8368 & \textit{0.3140} & 5.3384 & 8.8236 \\
Sundial & 0.8434 & 0.3555 & 5.4162 & 8.8337 \\
TiRex & 0.8293 & 0.3191 & 5.3070 & 8.7965 \\
TimesFM-1.0 & 0.8575 & 0.3311 & 5.4873 & 8.9984 \\
TimesFM-2.0 & 0.8452 & 0.3234 & 5.4150 & 8.9688 \\
TimesFM-2.5 & \textit{0.8268} & \textbf{0.3109} & \underline{5.2671} & 8.7433 \\
VisionTS++ & \textbf{0.8201} & \underline{0.3128} & \textbf{5.2316} & \underline{8.6742} \\
\midrule
\multicolumn{5}{l}{\textbf{ML Baselines}} \\
DLinear & 0.8430 & 0.3370 & 5.4519 & \textit{8.7244} \\
DeepAR & 0.8672 & 0.3161 & 5.4928 & 9.2380 \\
LightGBM & 0.9190 & 0.3347 & 5.7991 & 9.5348 \\
PatchTST & \underline{0.8264} & 0.3246 & \textit{5.2885} & \textbf{8.6473} \\
\midrule
\multicolumn{5}{l}{\textbf{Statistical Baselines}} \\
AutoETS & 0.9247 & 0.3824 & 6.0361 & 9.5329 \\
Seasonal Naive & 1.0795 & 0.4722 & 7.0526 & 11.4082 \\
\bottomrule
\end{tabular}
\end{small}
\end{table}
\begin{table}[H]
\caption{PM2.5 leaderboard --- EEA FR}
\centering
\begin{small}
\begin{tabular}{lrrrr}
\toprule
model & MASE & CRPS & MAE & RMSE \\
\midrule
\multicolumn{5}{l}{\textbf{TSFMs}} \\
Chronos-2 & 0.8328 & 0.3404 & 3.2385 & 5.4543 \\
Chronos-Bolt & 0.8415 & 0.3453 & 3.2647 & 5.5422 \\
Kairos & 0.9590 & 0.4093 & 3.8036 & 6.1093 \\
Moirai-1 & 0.8430 & 0.3508 & 3.3001 & 5.5336 \\
Moirai-2 & 0.8268 & \textit{0.3344} & 3.1962 & 5.3406 \\
Sundial & 0.8452 & 0.3819 & 3.2904 & 5.4066 \\
TiRex & 0.8207 & 0.3380 & 3.1867 & 5.3400 \\
TimesFM-1.0 & 0.8572 & 0.3593 & 3.3354 & 5.5124 \\
TimesFM-2.0 & 0.8389 & 0.3464 & 3.2634 & 5.4468 \\
TimesFM-2.5 & \textit{0.8195} & \underline{0.3331} & \textit{3.1692} & \textit{5.3046} \\
VisionTS++ & \textbf{0.8163} & 0.3358 & \underline{3.1556} & \underline{5.2660} \\
\midrule
\multicolumn{5}{l}{\textbf{ML Baselines}} \\
DLinear & 0.8386 & 0.3463 & 3.2626 & 5.3056 \\
DeepAR & 0.8494 & 0.3457 & 3.2760 & 5.4698 \\
LightGBM & 0.9460 & 0.3584 & 3.5776 & 5.9298 \\
PatchTST & \underline{0.8189} & \textbf{0.3304} & \textbf{3.1556} & \textbf{5.2533} \\
\midrule
\multicolumn{5}{l}{\textbf{Statistical Baselines}} \\
AutoETS & 0.9203 & 0.4077 & 3.6309 & 5.8095 \\
Seasonal Naive & 1.0877 & 0.5087 & 4.2827 & 6.8934 \\
\bottomrule
\end{tabular}
\end{small}
\end{table}
\begin{table}[H]
\caption{SO2 leaderboard --- EEA FR}
\centering
\begin{small}
\begin{tabular}{lrrrr}
\toprule
model & MASE & CRPS & MAE & RMSE \\
\midrule
\multicolumn{5}{l}{\textbf{TSFMs}} \\
Chronos-2 & 1.0841 & 0.6271 & 1.4486 & 5.4111 \\
Chronos-Bolt & 1.0895 & 0.6357 & 1.4504 & 5.5090 \\
Kairos & 1.1456 & 0.8756 & 1.5378 & 5.3951 \\
Moirai-1 & 1.0801 & \textit{0.5783} & 1.4103 & 5.2885 \\
Moirai-2 & 1.0719 & 0.5813 & 1.4000 & 5.2781 \\
Sundial & 1.0877 & 0.9501 & 1.5083 & \textbf{5.2055} \\
TiRex & \textbf{1.0594} & 0.5987 & \textbf{1.3797} & 5.2516 \\
TimesFM-1.0 & 1.0877 & 0.6174 & 1.4257 & 5.2673 \\
TimesFM-2.0 & 1.0798 & 0.6054 & 1.4012 & 5.2571 \\
TimesFM-2.5 & \textit{1.0687} & \underline{0.5512} & \underline{1.3831} & \textit{5.2422} \\
VisionTS++ & \underline{1.0627} & 0.6070 & 1.4137 & 5.2672 \\
\midrule
\multicolumn{5}{l}{\textbf{ML Baselines}} \\
DLinear & 1.1879 & 0.9633 & 1.6356 & \underline{5.2230} \\
DeepAR & 1.0962 & \textbf{0.5242} & 1.4141 & 5.3753 \\
LightGBM & 1.1133 & 0.8002 & 1.4508 & 5.2480 \\
PatchTST & 1.0905 & 0.6690 & \textit{1.3976} & 5.3111 \\
\midrule
\multicolumn{5}{l}{\textbf{Statistical Baselines}} \\
AutoETS & 1.2921 & 2.0221 & 2.1085 & 5.7417 \\
Seasonal Naive & 1.3894 & 2.6173 & 2.1497 & 7.1615 \\
\bottomrule
\end{tabular}
\end{small}
\end{table}

\FloatBarrier
\subsection{EPA}
\begin{table}[H]
\caption{CO leaderboard --- EPA}
\centering
\begin{small}
\begin{tabular}{lrrrr}
\toprule
model & MASE & CRPS & MAE & RMSE \\
\midrule
\multicolumn{5}{l}{\textbf{TSFMs}} \\
Chronos-2 & 0.8876 & 0.4234 & 0.0924 & 0.1540 \\
Chronos-Bolt & 0.8950 & 0.4590 & 0.0930 & 0.1551 \\
Kairos & 1.1074 & 0.5472 & 0.1186 & 0.1837 \\
Moirai-1 & 0.9165 & 0.4215 & 0.0955 & 0.1566 \\
Moirai-2 & 0.8918 & \textbf{0.4034} & 0.0927 & 0.1530 \\
Sundial & 0.9020 & 0.5943 & 0.0941 & 0.1518 \\
TiRex & \underline{0.8783} & 0.4389 & \textit{0.0918} & 0.1531 \\
TimesFM-1.0 & 0.9270 & 0.5742 & 0.0961 & 0.1566 \\
TimesFM-2.0 & 0.9012 & 0.5556 & 0.0935 & 0.1545 \\
TimesFM-2.5 & \textit{0.8840} & \textit{0.4194} & \underline{0.0916} & \underline{0.1513} \\
VisionTS++ & \textbf{0.8694} & 0.4490 & \textbf{0.0904} & \textbf{0.1501} \\
\midrule
\multicolumn{5}{l}{\textbf{ML Baselines}} \\
DLinear & 1.1380 & 0.6602 & 0.0958 & \textit{0.1515} \\
DeepAR & 1.3739 & \underline{0.4077} & 0.1003 & 0.1606 \\
LightGBM & 1.0428 & 0.5774 & 0.0965 & 0.1567 \\
PatchTST & 0.9606 & 0.4638 & 0.0936 & 0.1551 \\
\midrule
\multicolumn{5}{l}{\textbf{Statistical Baselines}} \\
AutoETS & 1.0302 & 0.6398 & 0.1094 & 0.1670 \\
Seasonal Naive & 1.1079 & 0.7980 & 0.1195 & 0.1996 \\
\bottomrule
\end{tabular}
\end{small}
\end{table}
\begin{table}[H]
\caption{NO2 leaderboard --- EPA}
\centering
\begin{small}
\begin{tabular}{lrrrr}
\toprule
model & MASE & CRPS & MAE & RMSE \\
\midrule
\multicolumn{5}{l}{\textbf{TSFMs}} \\
Chronos-2 & 0.8219 & 0.4729 & 6.3948 & 9.6644 \\
Chronos-Bolt & 0.8195 & 0.4806 & 6.3767 & 9.5964 \\
Kairos & 1.0933 & 0.6577 & 8.4913 & 11.8802 \\
Moirai-1 & 0.8351 & 0.5070 & 6.5003 & 9.7437 \\
Moirai-2 & 0.8207 & 0.4768 & 6.3987 & 9.6603 \\
Sundial & 0.8345 & 0.5839 & 6.5398 & 9.6261 \\
TiRex & \textit{0.8184} & \textit{0.4707} & 6.3913 & 9.6839 \\
TimesFM-1.0 & 0.8374 & 0.5149 & 6.5608 & 9.8290 \\
TimesFM-2.0 & 0.8250 & 0.4863 & 6.4367 & 9.7710 \\
TimesFM-2.5 & \underline{0.8134} & \underline{0.4632} & \underline{6.3313} & 9.5473 \\
VisionTS++ & \textbf{0.8028} & 0.4775 & \textbf{6.2476} & \underline{9.4596} \\
\midrule
\multicolumn{5}{l}{\textbf{ML Baselines}} \\
DLinear & 0.8330 & 0.5587 & 6.4987 & \textit{9.4786} \\
DeepAR & 0.8551 & \textbf{0.4241} & 6.5006 & 9.8773 \\
LightGBM & 0.8763 & 0.6080 & 6.8518 & 9.7412 \\
PatchTST & 0.8190 & 0.4999 & \textit{6.3553} & \textbf{9.3660} \\
\midrule
\multicolumn{5}{l}{\textbf{Statistical Baselines}} \\
AutoETS & 0.9506 & 0.6841 & 7.4490 & 10.4995 \\
Seasonal Naive & 1.0615 & 0.8330 & 8.4073 & 12.4848 \\
\bottomrule
\end{tabular}
\end{small}
\end{table}
\begin{table}[H]
\caption{Ozone leaderboard --- EPA}
\centering
\begin{small}
\begin{tabular}{lrrrr}
\toprule
model & MASE & CRPS & MAE & RMSE \\
\midrule
\multicolumn{5}{l}{\textbf{TSFMs}} \\
Chronos-2 & 0.8380 & 0.1851 & 12.2629 & 16.6584 \\
Chronos-Bolt & 0.8291 & \underline{0.1847} & 12.1424 & 16.4878 \\
Kairos & 1.4682 & 0.3221 & 21.3094 & 27.7594 \\
Moirai-1 & 0.8718 & 0.1984 & 12.7875 & 17.1063 \\
Moirai-2 & 0.8546 & 0.1941 & 12.5470 & 16.8949 \\
Sundial & 0.8285 & 0.2080 & 12.1178 & \textit{16.1548} \\
TiRex & \textbf{0.8184} & \textbf{0.1826} & \underline{11.9921} & 16.2554 \\
TimesFM-1.0 & 0.8909 & 0.2158 & 13.1105 & 17.5651 \\
TimesFM-2.0 & 0.8369 & 0.1949 & 12.2496 & 16.4241 \\
TimesFM-2.5 & \textit{0.8237} & \textit{0.1849} & \textit{12.0543} & 16.2240 \\
VisionTS++ & 0.8270 & 0.1863 & 12.1034 & 16.3623 \\
\midrule
\multicolumn{5}{l}{\textbf{ML Baselines}} \\
DLinear & 0.8399 & 0.2005 & 12.2227 & \underline{16.1133} \\
DeepAR & 0.8981 & 0.2354 & 13.0375 & 17.2487 \\
LightGBM & 0.8539 & 0.2097 & 12.4312 & 16.3009 \\
PatchTST & \underline{0.8208} & 0.1981 & \textbf{11.9470} & \textbf{15.7365} \\
\midrule
\multicolumn{5}{l}{\textbf{Statistical Baselines}} \\
AutoETS & 0.9526 & 0.2184 & 13.9780 & 18.1409 \\
Seasonal Naive & 1.0477 & 0.2486 & 15.4815 & 20.6695 \\
\bottomrule
\end{tabular}
\end{small}
\end{table}
\begin{table}[H]
\caption{PM10 leaderboard --- EPA}
\centering
\begin{small}
\begin{tabular}{lrrrr}
\toprule
model & MASE & CRPS & MAE & RMSE \\
\midrule
\multicolumn{5}{l}{\textbf{TSFMs}} \\
Chronos-2 & 0.8628 & 0.3777 & 10.1982 & 27.9851 \\
Chronos-Bolt & 0.8668 & 0.3813 & 10.3085 & 29.0075 \\
Kairos & 0.9902 & 0.4428 & 11.8381 & 29.2064 \\
Moirai-1 & 0.8765 & 0.3827 & 10.3722 & 28.0290 \\
Moirai-2 & 0.8642 & \textit{0.3708} & 10.1868 & 27.8411 \\
Sundial & 0.8736 & 0.4338 & 10.4963 & \textbf{27.5656} \\
TiRex & \underline{0.8556} & 0.3747 & \textit{10.0847} & \textit{27.6752} \\
TimesFM-1.0 & 0.8754 & 0.3802 & 10.3203 & 27.8795 \\
TimesFM-2.0 & 0.8712 & 0.3733 & 10.2584 & 27.9821 \\
TimesFM-2.5 & \textit{0.8567} & \textbf{0.3644} & \underline{10.0779} & 27.7150 \\
VisionTS++ & \textbf{0.8476} & \underline{0.3698} & \textbf{10.0027} & \underline{27.6283} \\
\midrule
\multicolumn{5}{l}{\textbf{ML Baselines}} \\
DLinear & 0.8938 & 0.4170 & 11.1507 & 28.2917 \\
DeepAR & 0.8966 & 0.3865 & 10.4667 & 28.1655 \\
LightGBM & 0.9009 & 0.3937 & 10.5341 & 28.1171 \\
PatchTST & 0.8570 & 0.3716 & 10.1313 & 27.8679 \\
\midrule
\multicolumn{5}{l}{\textbf{Statistical Baselines}} \\
AutoETS & 1.0000 & 0.5669 & 13.1126 & 30.6961 \\
Seasonal Naive & 1.1234 & 0.7049 & 14.0309 & 37.1597 \\
\bottomrule
\end{tabular}
\end{small}
\end{table}
\begin{table}[H]
\caption{PM2.5 leaderboard --- EPA}
\centering
\begin{small}
\begin{tabular}{lrrrr}
\toprule
model & MASE & CRPS & MAE & RMSE \\
\midrule
\multicolumn{5}{l}{\textbf{TSFMs}} \\
Chronos-2 & 0.8239 & 0.3574 & 3.0071 & 5.6841 \\
Chronos-Bolt & 0.8319 & 0.3623 & 3.0459 & 6.0134 \\
Kairos & 0.9258 & 0.4147 & 3.4086 & 5.9857 \\
Moirai-1 & 0.8321 & 0.3649 & 3.0339 & 5.6331 \\
Moirai-2 & 0.8164 & \underline{0.3494} & 2.9580 & 5.5068 \\
Sundial & 0.8258 & 0.3960 & 3.0149 & 5.4862 \\
TiRex & \underline{0.8071} & 0.3514 & \textit{2.9328} & 5.4541 \\
TimesFM-1.0 & 0.8379 & 0.3669 & 3.0485 & 5.5921 \\
TimesFM-2.0 & 0.8287 & 0.3609 & 3.0221 & 5.5883 \\
TimesFM-2.5 & \textit{0.8072} & \textbf{0.3462} & \underline{2.9242} & \textit{5.4421} \\
VisionTS++ & \textbf{0.8045} & \textit{0.3500} & \textbf{2.9182} & \textbf{5.4345} \\
\midrule
\multicolumn{5}{l}{\textbf{ML Baselines}} \\
DLinear & 0.8338 & 0.3634 & 3.0381 & 5.4803 \\
DeepAR & 0.8302 & 0.3587 & 3.0094 & 5.5858 \\
LightGBM & 0.8590 & 0.3886 & 3.1265 & 5.6449 \\
PatchTST & 0.8184 & 0.3634 & 2.9643 & \underline{5.4390} \\
\midrule
\multicolumn{5}{l}{\textbf{Statistical Baselines}} \\
AutoETS & 0.9187 & 0.4386 & 3.4324 & 6.0979 \\
Seasonal Naive & 1.0868 & 0.5546 & 4.0498 & 7.3536 \\
\bottomrule
\end{tabular}
\end{small}
\end{table}
\begin{table}[H]
\caption{SO2 leaderboard --- EPA}
\centering
\begin{small}
\begin{tabular}{lrrrr}
\toprule
model & MASE & CRPS & MAE & RMSE \\
\midrule
\multicolumn{5}{l}{\textbf{TSFMs}} \\
Chronos-2 & 1.2588 & 2.3582 & 1.3193 & 4.4735 \\
Chronos-Bolt & 1.2631 & 2.6726 & 1.3060 & 4.4660 \\
Kairos & 1.3632 & 4.2472 & 1.4510 & 4.5258 \\
Moirai-1 & 1.2703 & 2.1797 & 1.2996 & 4.4258 \\
Moirai-2 & 1.2544 & \textit{1.8346} & \textit{1.2766} & 4.3716 \\
Sundial & 1.2805 & 4.8651 & 1.3758 & \textbf{4.2742} \\
TiRex & \textbf{1.2411} & 2.2390 & \textbf{1.2516} & \underline{4.3317} \\
TimesFM-1.0 & 1.4186 & 2.9853 & 1.3194 & 4.3768 \\
TimesFM-2.0 & 1.3182 & 2.7053 & 1.2922 & 4.3542 \\
TimesFM-2.5 & \textit{1.2488} & 1.8708 & \underline{1.2691} & \textit{4.3334} \\
VisionTS++ & \underline{1.2463} & 2.5548 & 1.2969 & 4.3692 \\
\midrule
\multicolumn{5}{l}{\textbf{ML Baselines}} \\
DLinear & 1.4244 & 3.0468 & 1.5137 & 4.5582 \\
DeepAR & 1.3642 & \textbf{0.7531} & 1.3597 & 4.5083 \\
LightGBM & 1.3163 & 3.2585 & 1.3360 & 4.4334 \\
PatchTST & 1.3065 & \underline{1.2619} & 1.3155 & 4.5736 \\
\midrule
\multicolumn{5}{l}{\textbf{Statistical Baselines}} \\
AutoETS & 1.4663 & 11.4132 & 1.8435 & 4.6999 \\
Seasonal Naive & 1.5381 & 16.0548 & 1.8590 & 5.7495 \\
\bottomrule
\end{tabular}
\end{small}
\end{table}

\FloatBarrier
\subsection{SINAICA}
\begin{table}[H]
\caption{CO leaderboard --- SINAICA}
\centering
\begin{small}
\begin{tabular}{lrrrr}
\toprule
model & MASE & CRPS & MAE & RMSE \\
\midrule
\multicolumn{5}{l}{\textbf{TSFMs}} \\
Chronos-2 & 1.6459 & 0.1534 & 0.4190 & 11.3693 \\
Chronos-Bolt & 1.6518 & 0.9410 & 0.6650 & 14.8405 \\
Kairos & 2.0228 & 0.4077 & 0.5559 & 11.5254 \\
Moirai-1 & 1.6864 & 0.1517 & 0.4240 & 11.3715 \\
Moirai-2 & 1.6456 & \textit{0.1486} & 0.4164 & 11.3628 \\
Sundial & 1.6552 & 0.2156 & 0.4415 & \textbf{11.3511} \\
TiRex & \textit{1.6403} & 0.1628 & \textit{0.4151} & 11.3623 \\
TimesFM-1.0 & 1.6776 & 0.1766 & 0.4314 & 11.3675 \\
TimesFM-2.0 & 1.6555 & 0.1542 & 0.4232 & 11.3645 \\
TimesFM-2.5 & \underline{1.6387} & \textbf{0.1441} & \underline{0.4114} & \underline{11.3532} \\
VisionTS++ & \textbf{1.6296} & \underline{0.1480} & \textbf{0.4096} & \textit{11.3598} \\
\midrule
\multicolumn{5}{l}{\textbf{ML Baselines}} \\
DLinear & 1.7215 & 0.5229 & 0.6342 & 12.0907 \\
DeepAR & 1.8409 & 0.3589 & 0.5240 & 11.4274 \\
LightGBM & 1.7096 & 0.3526 & 0.5807 & 12.0757 \\
PatchTST & 1.6721 & 0.2506 & 0.4678 & 11.3776 \\
\midrule
\multicolumn{5}{l}{\textbf{Statistical Baselines}} \\
AutoETS & 1.7730 & 1.0845 & 0.7294 & 12.1412 \\
Seasonal Naive & 1.8727 & 1.3228 & 0.6921 & 15.9470 \\
\bottomrule
\end{tabular}
\end{small}
\end{table}
\begin{table}[H]
\caption{NO2 leaderboard --- SINAICA}
\centering
\begin{small}
\begin{tabular}{lrrrr}
\toprule
model & MASE & CRPS & MAE & RMSE \\
\midrule
\multicolumn{5}{l}{\textbf{TSFMs}} \\
Chronos-2 & 0.8690 & 0.2712 & 9.6182 & 14.7003 \\
Chronos-Bolt & \underline{0.8669} & 0.2734 & 9.5991 & 14.6379 \\
Kairos & 1.1896 & 0.3745 & 12.6191 & 17.8018 \\
Moirai-1 & 0.8812 & 0.2781 & 9.7046 & 14.6250 \\
Moirai-2 & 0.8685 & \textit{0.2704} & 9.5532 & 14.5262 \\
Sundial & 0.8800 & 0.2997 & 9.6786 & \textit{14.4054} \\
TiRex & 0.8723 & 0.2741 & 9.6286 & 14.6021 \\
TimesFM-1.0 & 0.8953 & 0.2864 & 9.8791 & 14.8711 \\
TimesFM-2.0 & 0.8718 & 0.2718 & \textit{9.5517} & 14.4977 \\
TimesFM-2.5 & \textit{0.8683} & \underline{0.2699} & \underline{9.5247} & 14.4222 \\
VisionTS++ & \textbf{0.8508} & \textbf{0.2660} & \textbf{9.3498} & \underline{14.2366} \\
\midrule
\multicolumn{5}{l}{\textbf{ML Baselines}} \\
DLinear & 0.9317 & 0.2989 & 10.2054 & 14.8442 \\
DeepAR & 0.9074 & 0.2959 & 9.8621 & 14.5161 \\
LightGBM & 0.9113 & 0.2846 & 9.9901 & 15.0076 \\
PatchTST & 0.8993 & 0.2882 & 9.7980 & \textbf{14.2044} \\
\midrule
\multicolumn{5}{l}{\textbf{Statistical Baselines}} \\
AutoETS & 0.9776 & 0.3215 & 10.8688 & 15.7756 \\
Seasonal Naive & 1.0652 & 0.3690 & 11.9852 & 18.1850 \\
\bottomrule
\end{tabular}
\end{small}
\end{table}
\begin{table}[H]
\caption{Ozone leaderboard --- SINAICA}
\centering
\begin{small}
\begin{tabular}{lrrrr}
\toprule
model & MASE & CRPS & MAE & RMSE \\
\midrule
\multicolumn{5}{l}{\textbf{TSFMs}} \\
Chronos-2 & \textit{0.9148} & 0.2596 & \textit{15.5821} & \underline{23.9814} \\
Chronos-Bolt & 0.9280 & 0.2616 & 15.8567 & 24.7865 \\
Kairos & 1.7499 & 0.4667 & 29.4913 & 42.0203 \\
Moirai-1 & 0.9664 & 0.2784 & 16.5288 & 25.2773 \\
Moirai-2 & 0.9261 & 0.2602 & 15.7702 & 24.2713 \\
Sundial & 0.9369 & 0.2897 & 16.0107 & 24.4242 \\
TiRex & 0.9205 & \textit{0.2590} & 15.7612 & 24.4908 \\
TimesFM-1.0 & 0.9849 & 0.2776 & 16.7733 & 25.6076 \\
TimesFM-2.0 & 0.9520 & 0.2708 & 16.2341 & 25.0203 \\
TimesFM-2.5 & 0.9214 & 0.2618 & 15.6947 & \textit{24.1613} \\
VisionTS++ & \underline{0.9126} & \textbf{0.2561} & \underline{15.5572} & 24.4462 \\
\midrule
\multicolumn{5}{l}{\textbf{ML Baselines}} \\
DLinear & 0.9437 & 0.2689 & 16.0446 & 24.4160 \\
DeepAR & 0.9585 & 0.2786 & 16.2756 & 24.9166 \\
LightGBM & 0.9979 & 0.2908 & 17.0597 & 25.4262 \\
PatchTST & \textbf{0.9095} & \underline{0.2589} & \textbf{15.4667} & \textbf{23.4825} \\
\midrule
\multicolumn{5}{l}{\textbf{Statistical Baselines}} \\
AutoETS & 1.0582 & 0.3151 & 18.1458 & 26.4147 \\
Seasonal Naive & 1.0679 & 0.3324 & 18.4489 & 28.8281 \\
\bottomrule
\end{tabular}
\end{small}
\end{table}
\begin{table}[H]
\caption{PM10 leaderboard --- SINAICA}
\centering
\begin{small}
\begin{tabular}{lrrrr}
\toprule
model & MASE & CRPS & MAE & RMSE \\
\midrule
\multicolumn{5}{l}{\textbf{TSFMs}} \\
Chronos-2 & \textit{0.8668} & 0.2898 & 21.1401 & 41.3729 \\
Chronos-Bolt & 0.8735 & 0.2956 & 21.3051 & 41.6046 \\
Kairos & 1.0590 & 0.3505 & 25.2880 & 44.6913 \\
Moirai-1 & 0.8914 & 0.2996 & 21.6874 & 40.9342 \\
Moirai-2 & 0.8698 & \textit{0.2873} & 21.1115 & \textit{40.3774} \\
Sundial & 0.8821 & 0.3284 & 21.6142 & 40.6053 \\
TiRex & 0.8700 & 0.2920 & \textit{21.0696} & 40.4695 \\
TimesFM-1.0 & 0.8875 & 0.2973 & 21.6327 & 41.1887 \\
TimesFM-2.0 & 0.8744 & 0.2930 & 21.2906 & 40.8761 \\
TimesFM-2.5 & \underline{0.8648} & \textbf{0.2844} & \underline{20.9067} & \underline{40.2202} \\
VisionTS++ & \textbf{0.8502} & \underline{0.2846} & \textbf{20.6710} & \textbf{40.0188} \\
\midrule
\multicolumn{5}{l}{\textbf{ML Baselines}} \\
DLinear & 0.8913 & 0.3118 & 22.1349 & 40.8321 \\
DeepAR & 0.8965 & 0.3126 & 22.2166 & 42.4870 \\
LightGBM & 0.9108 & 0.3215 & 22.7085 & 42.9164 \\
PatchTST & 0.8858 & 0.2934 & 21.7989 & 41.8590 \\
\midrule
\multicolumn{5}{l}{\textbf{Statistical Baselines}} \\
AutoETS & 0.9814 & 0.3616 & 24.6224 & 43.8215 \\
Seasonal Naive & 1.0909 & 0.4343 & 27.8025 & 53.0974 \\
\bottomrule
\end{tabular}
\end{small}
\end{table}
\begin{table}[H]
\caption{PM2.5 leaderboard --- SINAICA}
\centering
\begin{small}
%
\end{small}
\end{table}
\begin{table}[H]
\caption{SO2 leaderboard --- SINAICA}
\centering
\begin{small}
%
\end{small}
\end{table}

\onecolumn
\section{Dataset Statistics}
\label{app:data_statistics}
Table~\ref{tab:data_statistics} reports distributional statistics for each network-pollutant combination, averaged over sites.

\begin{small}
%
\end{small}

\clearpage

\section{Aggregated Results by Network and Pollutant}
\label{app:aggregated}

Table~\ref{tab:per_dataset} and Table~\ref{tab:per_pollutant} disaggregate the overall leaderboard by network and by pollutant respectively, both normalized by Seasonal Naive. These tables reveal where relative model rankings remain stable and where task difficulty varies most, complementing the per-pollutant detail in Appendix~\ref{app:per_pollutant}.

\begin{small}
%
\end{small}

\clearpage

\end{document}